\documentclass{article}

\usepackage[preprint]{neurips_2026}

\usepackage[utf8]{inputenc} 
\usepackage[T1]{fontenc}    
\usepackage{hyperref}       
\usepackage{url}            
\usepackage{booktabs}       
\usepackage{amsfonts}       
\usepackage{nicefrac}       
\usepackage{microtype}      
\usepackage{xcolor}         

\usepackage{graphicx}
\usepackage{subcaption}
\usepackage{multirow}
\usepackage{enumitem}

\usepackage{amsmath}
\usepackage{amssymb}
\usepackage{mathtools}
\usepackage{amsthm}
\usepackage{algorithm}
\usepackage{algorithmic}

\usepackage[capitalize,noabbrev]{cleveref}

\theoremstyle{plain}
\newtheorem{theorem}{Theorem}[section]
\newtheorem{proposition}[theorem]{Proposition}
\newtheorem{lemma}[theorem]{Lemma}
\newtheorem{corollary}[theorem]{Corollary}
\theoremstyle{definition}

\theoremstyle{remark}

\hypersetup{hidelinks}

\title{What are Key Factors for Updates in RL for LLM Reasoning?}

\author{%
  \textbf{Peidong Wang}\textsuperscript{1,2}\thanks{Equal contribution. Work done during internship at Microsoft Research Asia.}
  \quad
  \textbf{Demi Wang}\textsuperscript{2,3}\footnotemark[1]
  \quad
  \textbf{Xufang Luo}\textsuperscript{2}\thanks{Corresponding author: \texttt{luoxufang@outlook.com}.}
  \quad
  \textbf{Jiahang Xu}\textsuperscript{2}
  \\
  \textbf{Xiaocui Yang}\textsuperscript{1}
  \quad
  \textbf{Shi Feng}\textsuperscript{1}\thanks{Corresponding author: \texttt{fengshi@cse.neu.edu.cn}.}
  \quad
  \textbf{Yuqing Yang}\textsuperscript{2}
  \quad
  \textbf{Dongsheng Li}\textsuperscript{2}
  \\[0.5em]
  \normalfont\textsuperscript{1}Northeastern University
  \\
  \normalfont\textsuperscript{2}Microsoft Research
  \\
  \normalfont\textsuperscript{3}Carnegie Mellon University
}

\begin{document}

\maketitle

\begin{abstract}

Reinforcement Learning from Verifiable Rewards (RLVR) has emerged as a promising framework for enhancing the reasoning ability of large language models. However, much of the existing work is guided by heuristic intuition, leading to divergent algorithmic choices, even contradictory ones that nevertheless report empirical gains. To better understand this phenomenon, we conduct a theoretical analysis of RLVR updates. Our study reveals that differences in off-policy degree, determined by the number of gradient steps per rollout, substantially affect the distribution of importance sampling ratios and their clipping behavior, thereby altering which tokens dominate the update. Building on this insight, we characterize gradient expectation as the central quantity governing update dynamics and analyze the roles of token probability, advantage, and importance sampling ratio. Motivated by these findings, we propose Adaptive Clip Policy Optimization (ACPO), which adjusts clipping boundaries across token groups according to the empirical variance of their importance sampling ratios. Experiments on 3B and 7B models across diverse reasoning benchmarks, spanning mathematical problem solving, tabular QA, and logic puzzles, demonstrate that ACPO outperforms strong baselines such as DAPO and CISPO. These results demonstrate that principled, analysis-driven approaches yield more robust and effective RLVR methods. 
Code is available in: \url{https://github.com/Control-derek/ACPO}
\end{abstract}

\section{Introduction}
\label{sec:intro}

Reinforcement Learning from Verifiable Rewards (RLVR) aims to enhance the reasoning ability of large language models (LLMs) by optimizing against verifiable outcomes, such as mathematical correctness or logical validity \citep{shao2024deepseekmath, guo2025deepseek, lambert2024tulu3, jaech2024openai, chen2025acereason, yang2025qwen3}. This setting provides a scalable way to directly reward structured reasoning, making RLVR a central approach for advancing reliability in complex problem-solving tasks.

\begin{figure*}[!t]
  \centering
  \includegraphics[width=0.8\linewidth]{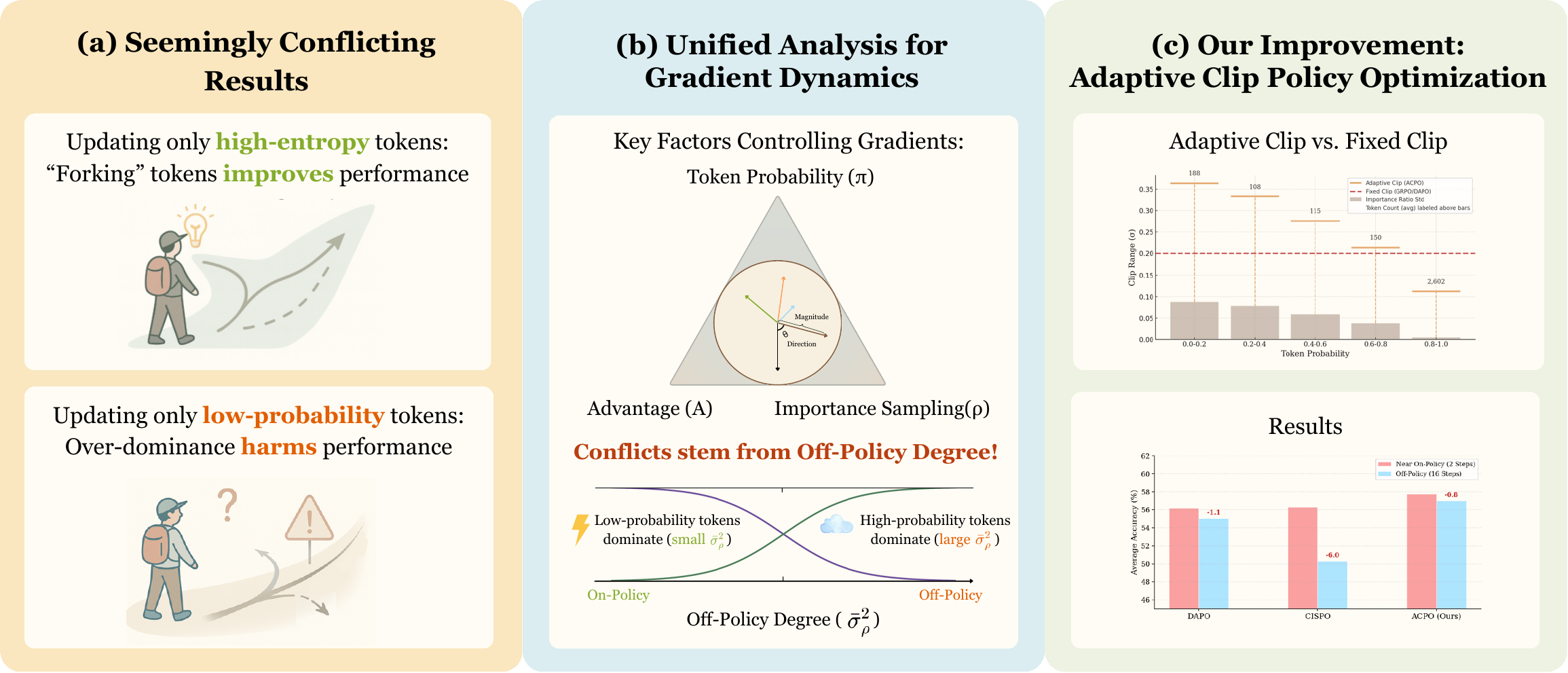}
  \caption{(a) Conflicting results: high-entropy updates help, while low-probability dominance hurts stability. (b) Our analysis attributes this to the clipped update, which shifts with off-policy degree via token probability, advantage, and IS. (c) ACPO adapts clipping per token group, improving over DAPO/CISPO across three reasoning tasks and off-policy regimes.}
  \label{fig:main}
  \vskip -0.2in
\end{figure*}

Despite substantial progress, RLVR research remains largely heuristic-driven rather than systematic analysis. Accordingly, many algorithmic variants have been proposed \citep{yu2025dapo, yue2025vapo, su2025klear, zheng2025group}, yet their effectiveness is still not well understood. As illustrated in Figure~\ref{fig:main}(a), even seemingly contradictory token-level heuristics report similar gains: \citet{wang2025beyond} prioritize high-entropy tokens and mask low-entropy ones, arguing that uncertain tokens are pivotal for reasoning, whereas \citet{yang2025not} emphasize high-probability tokens to prevent low-probability tokens from dominating gradient updates. Since token entropy is inversely correlated with token probability, these prescriptions are effectively opposite, yet both work in practice.

Motivated by the importance of token-level understanding in RLVR \citep{yang2025not, wang2025beyond, cui2025entropy, chen2025minimax, wang2025stabilizing}, we study the learning dynamics of clipped policy-gradient updates in RLVR.
Our starting point is a simple but under-emphasized observation: seemingly contradictory token-selection heuristics are often evaluated under \emph{different degrees of off-policy}.
For instance, the high-entropy update strategy uses many more updates per rollout than the high-probability one, causing the optimized policy to drift further from the behavior policy that generated the rollouts.
In practice, such off-policy regimes can be unavoidable because updates are often the bottleneck: one can cheaply generate many rollouts, but can only fit a limited number of update samples per step, making it natural to reuse each rollout across multiple updates.

Because GRPO~\citep{shao2024deepseekmath} relies on a clipped importance-sampling (IS) ratio, off-policy degree affects both token-gradient magnitude and \emph{which tokens survive clipping and contribute updates}.
This implies that analyses based on per-token intuition can be misleading: the overall update is an aggregation over tokens, and clipping can systematically suppress different token groups, changing both the update magnitude and direction.
Instead, one must examine the \textbf{expected effective gradient under clipping}.
In Section~\ref{sec:analysis_contradictions}, we quantify off-policy degree via the variance of the IS ratio and derive a closed-form characterization of the conditional gradient expectation.
A key consequence is a \textbf{gradient dominance reversal}: in near on-policy regimes, low-probability tokens can dominate the effective update, while in sufficiently off-policy regimes their IS ratios become highly dispersed and are frequently clipped, causing higher-probability tokens to dominate.
This reversal unifies the conflicting empirical findings in prior work.

Beyond resolving this contradiction, we ask more generally: what governs (i) which tokens contribute to the update, and (ii) how the update direction deviates from the on-policy ideal?
In Section~\ref{sec:factors}, we provide a component-wise analysis of RLVR updates along two dimensions: the \emph{magnitude} of the effective update and its \emph{directional bias} relative to the on-policy gradient. We show that token probability, advantage sign/magnitude, and IS-ratio clipping interact to shape both.
These results highlight a fundamental mismatch in standard practice: a single global clipping window is not aligned with the highly heterogeneous IS-ratio dispersion across token groups and training regimes, and can also substantially rotate the optimization direction.

Motivated by these insights, we propose \textbf{Adaptive Clip Policy Optimization (ACPO)} in Section~\ref{sec:acpo}.
ACPO replaces the fixed global clipping range with bin-wise thresholds that adapt to heterogeneous IS-ratio dispersion across token groups, reducing uneven clipping and mitigating clipping-induced gradient bias.
Empirically, ACPO consistently outperforms strong baselines (e.g., DAPO~\citep{yu2025dapo} and CISPO~\citep{chen2025minimax}) across three reasoning tasks under both near on-policy and off-policy training regimes. In summary, our main contributions are as follows:
\begin{itemize}[leftmargin=*,noitemsep,topsep=2pt,parsep=0pt,partopsep=0pt]
  \item Analysis of clipped RLVR: off-policy degree shifts gradient expectation via IS-ratio dispersion, causing gradient dominance reversal and reconciling prior heuristics.
  \item Joint effects of token probability, advantage, and IS-ratio clipping on update magnitude/direction, explaining why a single global clip can be mismatched.
  \item ACPO: variance-aware adaptive clipping with consistent gains over strong baselines across model scales, tasks, and both near on-policy and off-policy regimes.
\end{itemize}

\section{Theoretical Analysis of Empirical Contradictions}
\label{sec:analysis_contradictions}

\textbf{Preliminary.}
\label{sec:preliminaries}
In contrast to traditional actor-critic algorithms such as Proximal Policy Optimization (PPO) \citep{schulman2017proximal}, Group Relative Policy Optimization (GRPO) \citep{shao2024deepseekmath} removes the critic (of comparable size to the policy) and estimates advantages by standardizing rewards within a sampled group. For each prompt $q \sim P(Q)$, the policy \(\pi_{\theta}\) samples \(G\) responses \(\{o_i\}_{i=1}^G\), each scored by a rule-based reward \(r_i = R(o_i)\). The relative advantage is \(A_i = \frac{r_i - \operatorname{mean}(\{r_j\}_{j=1}^G)}{ \operatorname{std}(\{r_j\}_{j=1}^G)}\), where \(\operatorname{mean}\) and \(\operatorname{std}\) denote the sample mean and standard deviation, respectively. The objective is:

\vskip -0.2in
\begin{equation}
\small
\begin{split}
J_{GRPO}(\theta) = \mathbb{E}_{q \sim P(Q), \{o_i\}_{i=1}^G \sim \pi_{\theta_{old}}(O|q)} \frac{1}{G}\sum_{i=1}^{G}\frac{1}{|o_i|}\sum_{t=1}^{|o_i|} \{\min[\rho_{i,t}A_i, \operatorname{clip}(\rho_{i,t}, 1-\epsilon_l, 1+\epsilon_h)A_i]\}
\end{split}
\label{eq:GRPO}
\end{equation}

\vskip -0.2in

where $\rho_{i,t}=\frac{\pi_{\theta}(o_{i,t}|q,o_{i,<t})}{\pi_{\theta_{old}}(o_{i,t}|q,o_{i,<t})}$ is the importance sampling (IS) ratio, used to correct for off-policy updates. The clipping function $clip(\cdot, 1-\epsilon_l, 1+\epsilon_h)$ restricts this ratio to $[1 - \epsilon_l, 1 + \epsilon_h]$ for stable updates. GRPO typically uses symmetric clipping (\(\epsilon_l=\epsilon_h\)), while DAPO \citep{yu2025dapo} adopts asymmetric clipping (\(\epsilon_h>\epsilon_l\)) to mitigate entropy collapse. Following recent practice \citep{yu2025dapo,su2025klear,chen2025minimax}, we omit the KL penalty.

\subsection{Empirical Contradictions and the Role of Off-Policy Degree}
\label{sec:explain_the_difference}

\textbf{Contradictory Empirical Results.}
\label{sec:contradictions}
The limitations of a heuristic-driven approach are highlighted by recent, conflicting findings. \citet{wang2025beyond} report faster convergence and superior accuracy by selectively updating high-entropy tokens, as these tokens are pivotal for reasoning. In contrast, \citet{yang2025not} argue against amplifying the influence of low-probability tokens, showing that exclusively updating them leads to slow convergence and suboptimal performance. Their work suggests that the large, high-variance gradients from these tokens can destabilize training. This creates a paradox, as high-entropy tokens are often those with low generation probabilities \citep{guo2025segment, zheng2025first}. That two strategies targeting overlapping token populations yield divergent outcomes suggests that a critical factor is being overlooked. This puzzling discrepancy motivates our investigation into the fundamental structure of the policy gradient.

\textbf{Off-policy degree as the key.}
The key experimental difference between~\citet{yang2025not,wang2025beyond} is the number of updates per rollout: $2$ vs.
$16$.
More updates push the current policy further away from the behavior policy that generated the rollouts, increasing the degree of off-policy.
Because Eq.~\eqref{eq:GRPO} uses a clipped IS ratio, off-policy degree affects not only the magnitude of token gradients but also which tokens survive clipping and contribute any update.
Therefore, we analyze the \textbf{expected effective gradient under clipping}.

To quantify off-policy degree mechanistically, we track how dispersed the IS ratio is.
Specifically, we model the conditional IS-ratio variance given the behavior-policy probability,
$\sigma_{\rho}^{2}(\pi)=\mathrm{Var}[\rho_{i,t}\mid\pi_{i,t}=\pi]$,
and later aggregate it into a scalar proxy for the overall off-policy degree.

\subsection{Detailed Analysis}

\textbf{Overview.}
We characterize which tokens dominate the effective GRPO update after clipping by studying a token-level update coefficient and its conditional expectation given the token's behavior-policy probability.
Let
\(
\pi_{i,t} \triangleq \pi_{\theta_{\text{old}}}(o_{i,t}\mid q,o_{i,<t})
\)
for token position \(t\) in sampled sequence \(i\).
We show that increasing off-policy-ness (e.g., more updates per rollout) can reverse which probability region dominates the effective gradient, yielding a \textbf{Gradient Dominance Reversal} that reconciles~\citet{yang2025not,wang2025beyond}.


\textbf{Gradient of GRPO.}
To connect our analysis to the GRPO objective, we start from its gradient and express it as a sum of token-level contributions:
\begin{equation}
\small
\begin{split}
\nabla_\theta J_{\text{GRPO}}(\theta)
&= \mathbb{E}_{q \sim P(Q), \{o_i\}_{i=1}^G \sim \pi_{\theta_{\text{old}}}(O\mid q)}
\frac{1}{G}\sum_{i=1}^{G}\frac{1}{|o_i|}\sum_{t=1}^{|o_i|}
g_{i,t}(\theta),\\
g_{i,t}(\theta)
&\triangleq \rho_{i,t} A_i \cdot \mathbb{I}_{\text{clip}}\!\left(\rho_{i,t}, A_i\right)\cdot
\nabla_\theta \log \pi_\theta(o_{i,t}\mid q,o_{i,<t}),
\end{split}
\label{eq:gradient}
\end{equation}
The clipping indicator is
\begin{equation}
\mathbb{I}_{\text{clip}}\!\left( \rho_{i,t}, A_i \right) =
\begin{cases}
0 & A_i > 0,\ \rho_{i,t} > 1 + \epsilon_h, \\
0 & A_i < 0,\ \rho_{i,t} < 1 - \epsilon_l, \\
1 & \text{otherwise.}
\end{cases}
\label{eq:indicator_clip}
\end{equation}

\textbf{From parameter gradients to an analyzable logit coefficient.}
Directly analyzing $g_{i,t}(\theta)$ at the parameter level is difficult because it entangles model Jacobians, so we instead analyze the initial update signal at the pre-softmax logit of the sampled token, where $\partial\log\pi_\theta(o_{i,t})/\partial z_{i,t,o_{i,t}} = 1-\pi_\theta(o_{i,t})$. Under the standard small-step regime, the resulting token-wise effective logit-gradient coefficient can be approximated as

\vskip -0.1in

\begin{equation}
G_{i,t} \triangleq (1-\pi_{i,t})\cdot \rho_{i,t}\cdot A_i \cdot \mathbb{I}_{\text{clip}}(\rho_{i,t},A_i).
\label{eq:sim_grad}
\end{equation}
Here we use \(\pi\in(0,1)\) to denote a specific value of random variable \(\pi_{i,t}\) when conditioning.
We study the conditional expectation \(\mathbb{E}[G_{i,t}\mid \pi_{i,t}=\pi]\), abbreviated as \(\mathbb{E}[G\mid \pi]\). The expectation is over the sampled prompts (and the induced randomness in \(A_i\) and \(\rho_{i,t}\)), conditioned on \(\pi_{i,t}=\pi\).

In Eq.~\ref{eq:sim_grad}, a token contributes only if \(\rho_{i,t}\) lies in the clipping interval \([1-\epsilon_l,\,1+\epsilon_h]\) (with the boundary chosen based on the sign of \(A_i\)).
Thus, analyzing \(\mathbb{E}[G_{i,t}\mid \pi_{i,t}=\pi]\) reduces to characterizing how often \(\rho_{i,t}\) falls inside this interval under the conditioning.

Conditioned on \(\pi_{i,t}=\pi\), \(\rho_{i,t}\) remains random because the current-policy probability \(\pi_{\theta}(o_{i,t}\mid q,o_{i,<t})\) varies across prompts and evolves across updates. 
We denote this variability by the conditional distribution \(p(\rho\mid \pi)\) and summarize its spread by the conditional variance
\begin{equation}
\sigma_{\rho}^{2}(\pi)\ \triangleq\ \mathrm{Var}\!\left[\rho_{i,t}\mid \pi_{i,t}=\pi\right].
\label{eq:var_rho}
\end{equation}

We treat $\sigma_{\rho}^{2}(\pi)$ as the local, token-group-specific measure of off-policy degree, and aggregate it across the rollout token-probability distribution $p(\pi)$ to obtain a single scalar measure of overall off-policy strength,
\begin{equation}
\bar\sigma_{\rho}^{2}\ \triangleq\ \mathbb{E}_{\pi\sim p(\pi)}\!\left[\sigma_{\rho}^{2}(\pi)\right].
\label{eq:var_rho_bar}
\end{equation}
As training becomes more off-policy, \(p(\rho\mid \pi)\) becomes more dispersed and both \(\sigma_{\rho}^{2}(\pi)\) and \(\bar\sigma_{\rho}^{2}\) grow, which changes the expected amount of clipping and thus reshapes \(\mathbb{E}[G\mid \pi]\).
Throughout, we use $\sigma_{\rho}^{2}(\pi)$ as the local axis and $\bar\sigma_{\rho}^{2}$ as the global axis of off-policy degree.

The key is to characterize how $\sigma_\rho^2(\pi)$ depends on $\pi$ under a small update step.
Using a first-order approximation (Appendix~\ref{app:derivation_lemma}), we obtain:

\begin{lemma}[Variance of Importance Sampling Ratio]\label{lemma:is_ratio_variance}
Under a small logit perturbation $\Delta y$ from one update step, for a token with probability $\pi$, the IS-ratio variance admits the approximation
\begin{equation}
    \sigma_\rho^2(\pi) = \kappa^2 (1-\pi)^2 + O(\kappa^3),
\label{eq:var_is_ratio}
\end{equation}
where $\kappa$ is an effective coefficient that summarizes the per-step logit-perturbation scale together with a vocabulary shape factor (see Appendix~\ref{app:derivation_lemma}).
The higher-order term $O(\kappa^3)$ is negligible under the standard small-step assumption.
\end{lemma}

\begin{figure}[t]
    \centering
    \begin{subfigure}{0.42\linewidth}
        \centering
        \includegraphics[width=\textwidth]{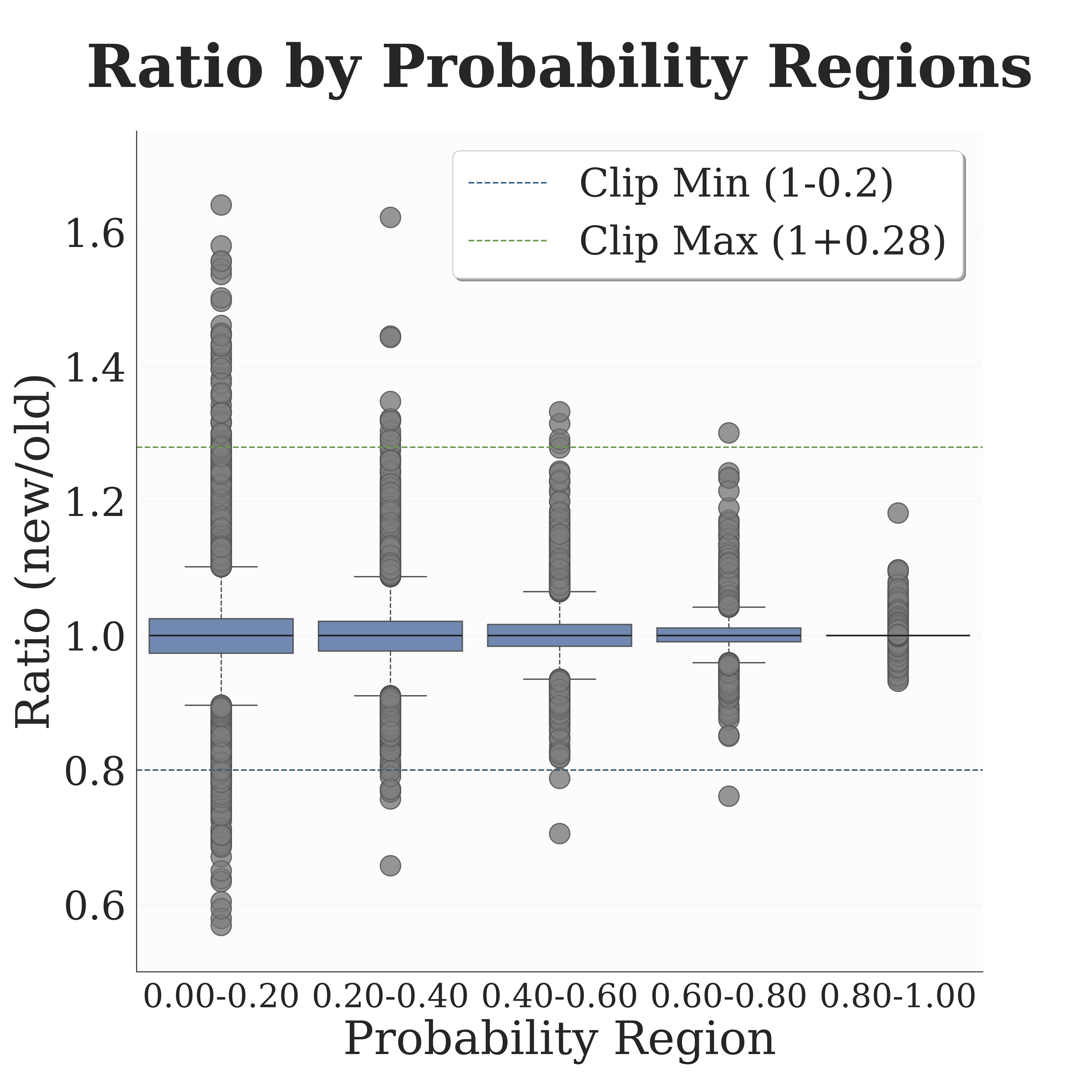}
        \label{fig:is_ratio_variance}
    \end{subfigure}
    \hfill
    \begin{subfigure}{0.42\linewidth}
        \centering
        \includegraphics[width=\textwidth]{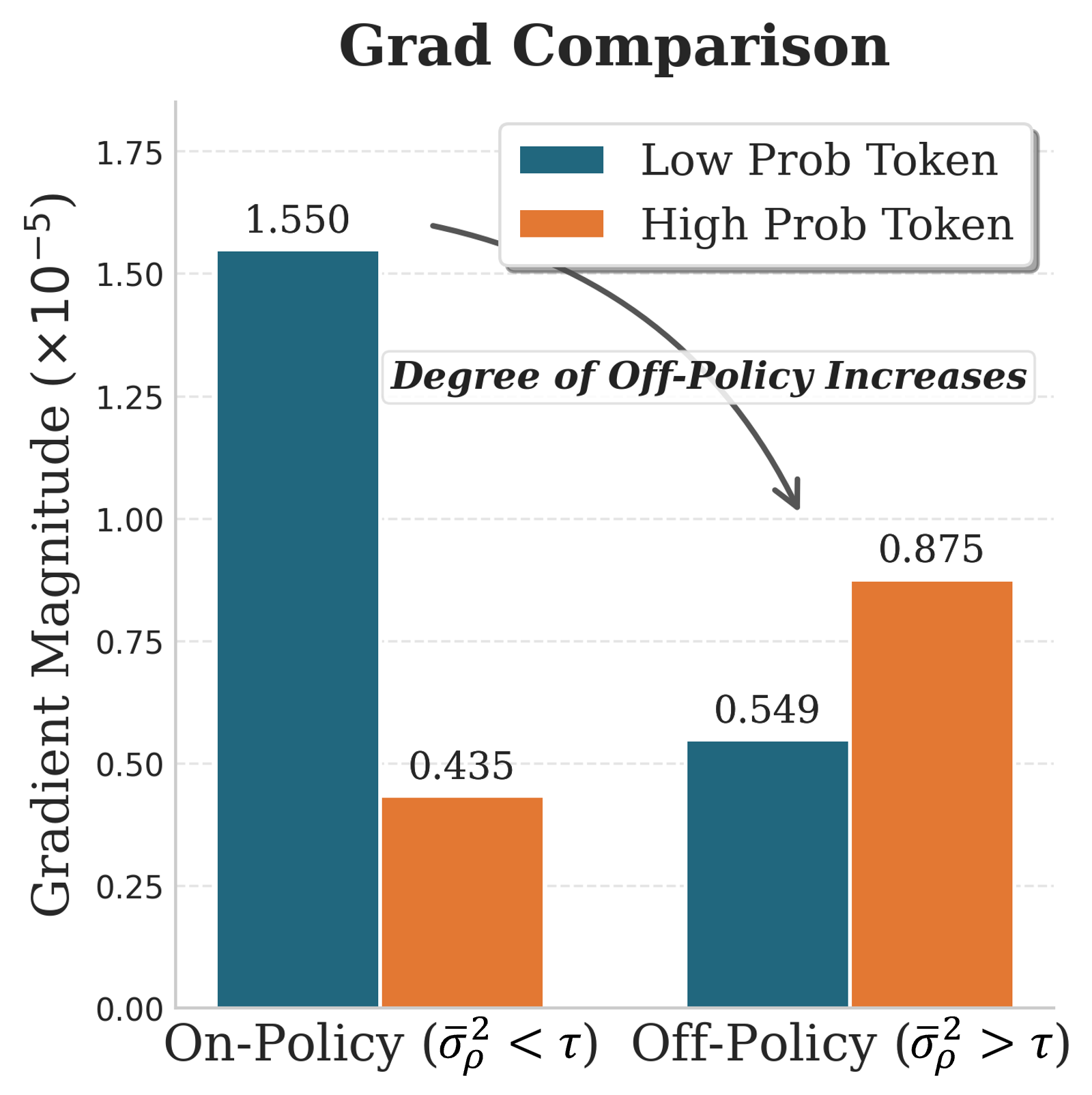}
        \label{fig:gradient_reversal}
    \end{subfigure}
    \caption{Empirical patterns of importance sampling ratio variance and gradient dominance. (a) Importance sampling ratios show higher variance for low-probability tokens. (b) As off-policy degree increases, gradient dominance shifts from low- to high-probability tokens.}
    \label{fig:empirical_validations_combined}
    \vskip -0.2in
\end{figure}

Eq.~\eqref{eq:var_is_ratio} shows that $\sigma_\rho^2(\pi)$ grows rapidly as $\pi$ decreases, implying a much wider IS-ratio distribution for low-probability tokens and hence a higher likelihood of being clipped. 
Figure~\ref{fig:empirical_validations_combined}(a) provides an empirical visualization of this principle. The distribution of the IS ratio for low-probability tokens is shown to have a much larger spread and numerous outliers compared to that of high-probability tokens.
We further fit $\mathrm{Var}(\rho)$ against $(1-\pi)$ and obtain an exponent of $1.90\pm 0.14$, closely matching the theoretical value of 2 (Appendix~Figure~\ref{fig:var_vs_prob_single}).

\textbf{Expectation of Gradient.}
Lemma~\ref{lemma:is_ratio_variance} characterizes how the local off-policy degree $\sigma_\rho^2(\pi)$ depends on $\pi$. We next translate this into an analytical approximation for $\mathbb{E}[G\mid \pi]$ under three modeling choices: (i) a log-normal IS ratio, $\log\rho\mid\pi\sim\mathcal{N}\!\big(-\tau^2(\pi)/2,\,\tau^2(\pi)\big)$, with $\tau(\pi):=\sqrt{\mathrm{Var}(\log\rho\mid\pi)}$ and the monotone reparametrization $\tau^2(\pi)=\log(1+\sigma_\rho^2(\pi))$; (ii) a linear coupling between conditional advantage and the centered log-ratio with $\pi$-dependent slope $\mu(\pi)\ge 0$; and (iii) the noiseless limit of this coupling for tractability. Full statements, finite-noise discussion, and a Gaussian re-derivation of the same qualitative conclusions are given in Appendix~\ref{app:derivation_proposition} and Appendix~\ref{app:gaussian_robustness}. Under these modeling choices, Appendix~\ref{app:derivation_proposition} derives the following expression.

\begin{proposition}[Conditional Gradient Expectation]\label{prop:gradient_expectation}
Fix a token with behavior-policy probability $\pi\in(0,1)$ and write $\tau=\tau(\pi)$.
Define the standardized clip endpoints
\begin{equation}
L(\pi)=\frac{\log(1-\epsilon_l)+\tau^2/2}{\tau},\qquad U(\pi)=\frac{\log(1+\epsilon_h)+\tau^2/2}{\tau},\qquad L'(\pi)=\min\big(L(\pi),0\big).
\end{equation}
The conditional gradient expectation admits the closed form
\begin{equation}
\mathbb{E}[G\mid\pi]\;=\;(1-\pi)\,\mu(\pi)\,H\!\big(\tau(\pi)\big),
\end{equation}
where
\begin{equation}
H(\tau)\;=\;\tau\Big\{\phi\!\big(L'-\tau\big)-\phi\!\big(U-\tau\big)\;+\;\tau\big[\Phi(U-\tau)-\Phi(L'-\tau)\big]\Big\}.
\end{equation}
\end{proposition}

The factor $(1-\pi)$ recovers the familiar on-policy logit-gradient front factor; $\mu(\pi)$ encodes the local coupling between advantage and the log IS-ratio; and $H(\tau)$ captures how the trust-region indicator selects which tokens contribute as the local off-policy degree grows. As $\tau\to 0$ (equivalently, $\sigma_\rho^2(\pi)\to 0$), $H(\tau)=\tau^2+O(\tau^4)=\sigma_\rho^2(\pi)+O\!\big(\sigma_\rho^4(\pi)\big)$, so the gradient signal scales as $\mu(\pi)(1-\pi)\,\sigma_\rho^2(\pi)$ in the near-on-policy regime. As $\sigma_\rho^2(\pi)$ grows, the trust-region window $[L'(\pi),U(\pi)]$ moves into the tail of the tilted log-normal density, the indicator mass $\Phi(U-\tau)-\Phi(L'-\tau)$ shrinks, and $H(\tau)$ eventually decreases, suppressing the contribution of tokens with high local off-policy degree (typically low-probability tokens, by Lemma~\ref{lemma:is_ratio_variance}).

\textbf{Two Different Gradient Dominances.}
Proposition~\ref{prop:gradient_expectation} enables us to compare the expected update contributions from different token populations under clipping. Partition the token-probability space into a low-probability interval $I_L=[0,p_L]$ and a high-probability interval $I_H=[p_H,1)$, with $0<p_L<p_H<1$, and let $\bar G_L(\bar\sigma_\rho^2)$ and $\bar G_H(\bar\sigma_\rho^2)$ denote the conditional averages of $\mathbb{E}[G\mid\pi]$ over $p(\pi)$ within $I_L$ and $I_H$, respectively (full definitions and derivations in Appendix~\ref{app:derivation_corollary}). Both depend on the global off-policy degree $\bar\sigma_\rho^2$ through the local axis $\sigma_\rho^2(\pi)$ entering $\mathbb{E}[G\mid\pi]$. We now characterize how the dominant contribution shifts as $\bar\sigma_\rho^2$ grows.



\begin{corollary}[Sufficient Condition for Gradient Dominance Reversal]\label{cor:gradient_dominance_reversal}
Under the modeling choices in Proposition~\ref{prop:gradient_expectation}, define $D(\bar\sigma_\rho^2):=\bar G_L(\bar\sigma_\rho^2)-\bar G_H(\bar\sigma_\rho^2)$.
Suppose
\textbf{(C1)} $\lim_{\bar\sigma_\rho^2\to 0^+} D(\bar\sigma_\rho^2)/\bar\sigma_\rho^2 = C_0 > 0$, where $C_0$ is a structural constant determined by $(\mu,p,I_L,I_H)$ (closed form in Appendix~\ref{app:derivation_corollary}); and
\textbf{(C2)} there exists $s_1>0$ with $D(s_1)<0$.
Then there exists $s_a\in(0,s_1)$ with $D(s_a)>0$, and by the intermediate value theorem at least one crossing point $s^\star\in(s_a,s_1)$ satisfies $D(s^\star)=0$. We refer to such $s^\star$ as a \textbf{gradient dominance reversal threshold}: when $D$ is sign-changing at $s^\star$, low-probability tokens dominate the conditional gradient signal in some left neighborhood of $s^\star$ and high-probability tokens dominate in some right neighborhood.
\end{corollary}

\emph{Remark.} (C1) and (C2) are concrete properties of $D$ that are verifiable from the closed form once $p(\pi)$ and $\mu(\pi)$ are specified; we verify both for the representative setup of this paper in Appendix~\ref{app:derivation_corollary}.

Fig.~\ref{fig:empirical_validations_combined}(b) confirms this prediction empirically: tracking gradient norms of Qwen2.5-7B~\citep{qwen2.5} trained on math reasoning with GRPO, low-probability tokens yield much larger gradients than high-probability tokens near on-policy (left), and as the off-policy degree increases (right) low-probability contributions collapse under clipping while high-probability tokens dominate.

\textbf{Resolving the Empirical Contradiction.}
This dominance reversal unifies the conflicting observations: \citet{yang2025not} operate at small $\bar\sigma_\rho^2$ where low-probability tokens dominate the gradient but inject high variance, so suppressing them stabilizes learning; \citet{wang2025beyond} operate at larger $\bar\sigma_\rho^2$ where low-probability tokens are frequently clipped (Figure~\ref{fig:empirical_validations_combined}(a)), so the surviving high-entropy contributors are comparatively stable and informative.

\section{Key Factors in LLM RL Updates}
\label{sec:factors}


Section~\ref{sec:analysis_contradictions} shows that seemingly conflicting token-update heuristics can be reconciled once we analyze the effective update under clipped importance sampling: the off-policy degree changes the IS-ratio dispersion, which in turn changes which tokens survive clipping and dominate the expected gradient. 
To turn this insight into principled algorithm design, we further ask: beyond token probability, what factors control (i) which tokens contribute to the update, and (ii) how the update direction deviates from the on-policy ideal?
This section provides a component-wise analysis of the GRPO update that will directly motivate adaptive clipping in Section~\ref{sec:acpo}.

\subsection{Analytical Framework: magnitude vs. direction}
\label{sec:framework}

We analyze GRPO updates from two perspectives.

\textbf{Magnitude (who dominates the update).}
Following Section~\ref{sec:analysis_contradictions}, we use the conditional gradient expectation $\mathbb{E}[G \mid \pi]$ (Proposition~\ref{prop:gradient_expectation}) to characterize \emph{effective} token contributions after clipping, capturing both per-token gradient scaling and clipping probability.

\textbf{Direction (how the optimization trajectory changes).}
Clipping is not merely a variance-reduction device: by zeroing out subsets of token-level updates, it can change the composition of the gradient and thus rotate the optimization direction even when the overall norm is similar. To quantify such directional changes in high dimensions, we adopt a layer-wise geometric measure: principal angles between the top-$k$ right-singular subspaces of two layer-gradient matrices, reporting mean angles over the top-$k$ directions (full procedure in Appendix~\ref{app:geom_details}).

\subsection{Key Factors}

Eq.~\ref{eq:sim_grad} shows that GRPO's \emph{effective} gradient is jointly determined by (i) token-wise sensitivity and probability-dependent IS variability, (ii) the advantage magnitude and sign, and (iii) IS-ratio clipping that gates off-policy contributions. 
We analyze each factor through both gradient expectation $\mathbb{E}[G \mid \pi]$ (magnitude dominance) and subspace alignment (directional distortion).


\textbf{Token properties: tail tokens are strong but fragile contributors.}
Low-probability tokens have large base sensitivity via $(1-\pi)$, making them potentially dominant contributors in near on-policy regimes (Corollary~\ref{cor:gradient_dominance_reversal}).
However, Lemma~\ref{lemma:is_ratio_variance} shows that their IS-ratio variance grows rapidly as $\pi$ decreases, increasing their clipping probability as training becomes more off-policy.
Empirically, gradients restricted to low-probability tokens exhibit disproportionately large norms and strong alignment with the full update (Appendix~\ref{app:token_analysis}), confirming that the learning signal is concentrated in the distribution tail, but that this signal is also the most susceptible to being filtered out by clipping.
This clarifies why entropy-/probability-based heuristics can both help: they are implicitly trading off signal strength (tail tokens) against stability under clipping.


\textbf{Advantage sign: negative-advantage samples steer the update direction.}
While advantage magnitude scales the update, its sign changes the qualitative effect on the policy.
Positive-advantage updates ($A>0$) reinforce sampled actions, whereas negative-advantage updates ($A<0$) penalize specific actions and effectively redistribute probability mass across many alternatives.
As a result, $A<0$ samples can dominate the directional structure even when the norms are comparable. 
Consistent with this mechanism, we find that the negative-advantage gradient subspace aligns substantially better with the final update direction than the positive-advantage one (Appendix~\ref{app:advantage_analysis}). 
This highlights that RLVR optimization is often driven more by error correction than by simply reinforcing already-good trajectories, and it implies that any gating mechanism (e.g., clipping) that disproportionately suppresses $A<0$ updates can significantly rotate the learning direction.

\textbf{Importance sampling and clipping: stability comes with directional distortion.}
Importance sampling enables reuse of rollouts under policy drift, but the variance of $\rho$ necessitates clipping for stable training.
Crucially, clipping gates token updates through $\mathbb{I}_{\text{clip}}(\rho, A)$, which changes the effective sample set and can therefore distort the update direction.
We empirically compare an on-policy gradient (aggregating samples into a single step) with the standard off-policy clipped gradient and observe a large directional divergence ($\approx 47^\circ$) despite similar magnitudes (Appendix~\ref{app:is_analysis}). 
This indicates that clipping does more than control variance: by down-weighting or discarding tokens with extreme IS ratios, it changes which tokens drive learning and can therefore shift the update direction away from the on-policy gradient, even when the gradient norm is similar.




\subsection{Summary and design implication}
\label{sec:takeaways}

Our analysis yields two algorithm design insights.

\textbf{Uniform clipping is mismatched to heterogeneous IS-ratio dispersion.}
Since IS-ratio variance depends strongly on token probability (Lemma~\ref{lemma:is_ratio_variance}) and changes with the off-policy degree, a single global clipping window makes the probability of being clipped highly uneven across token groups, systematically over-clipping some groups (typically low-probability tokens) while letting others dominate the effective update.

\textbf{Clipping affects not only variance but also the optimization trajectory.}
By changing which token updates survive, clipping can substantially rotate the gradient direction relative to the on-policy ideal, even when the overall gradient norm is preserved. This suggests that improving RLVR updates is not only about stabilizing magnitudes, but also about controlling which tokens steer the direction.

\textbf{Implication: clipping should adapt to the local policy shift.}
Together, these results suggest that the clipping window should be calibrated per token group according to its off-policy degree, rather than being fixed globally. 
A natural, directly observable proxy is the within-group IS-ratio variance, $\sigma_{\rho}^2(\pi)$.
This motivates the adaptive, variance-aware clipping strategy introduced in Section~\ref{sec:acpo}.

\section{Adaptive Clip Policy Optimization}
\label{sec:acpo}

\subsection{Motivation}



GRPO reuses rollout trajectories via importance-sampling (IS) ratios and stabilizes learning with clipping. The gradient of the clipped objective (Eq.~\ref{eq:gradient}, cf.~Eq.~\ref{eq:sim_grad}) implies that a token contributes to the update only when its IS ratio lies within the clipping window, so learning depends critically on how dispersed IS ratios are across tokens. Lemma~\ref{lemma:is_ratio_variance} provides an expression for the IS-ratio variance, implying substantially larger dispersion for low-probability tokens. With a single global clipping range, this leads to uneven clipping rates across token groups. Proposition~\ref{prop:gradient_expectation} and Corollary~\ref{cor:gradient_dominance_reversal} further show that such uneven clipping biases the expected gradient and can even change which token groups dominate the update, consistent with the update-direction distortions observed in Section~\ref{sec:factors}. 

\subsection{Adaptive Clip Policy Optimization (ACPO)}
\label{sec:acpo_section}

ACPO replaces the single global clipping range in Eq.~\ref{eq:GRPO} with a bin-specific range determined by the behavior-token probability. 
Given an update mini-batch (pooling all tokens in the update batch), we assign each token \((i,t)\) to one of \(B\) equal-width bins on \([0,1]\) via
\begin{equation}
\mathcal{B}(i,t)=\min\{B,\lfloor B\cdot \pi_{\theta_{\mathrm{old}}}(o_{i,t}\mid q,o_{i,<t})\rfloor+1\}.
\end{equation}
For each bin \(b\), we compute the within-bin IS-ratio dispersion and set the clipping range as
\begin{equation}
\sigma_b = \mathrm{Std}\big(\{\rho_{i,t}: \mathcal{B}(i,t)=b\}\big),\qquad
\epsilon_b = \epsilon_{\mathrm{base}}+\alpha\,\sigma_b.
\end{equation}
We apply token-wise clipping by using \(\epsilon_{\mathcal{B}(i,t)}\) in Eq.~\ref{eq:GRPO}. The objective is
\begin{equation}
\small
\begin{aligned}
J_{\text{ACPO}}(\theta) = \mathbb{E}_{q \sim P(Q),\, \{o_i\}_{i=1}^G \sim \pi_{\theta_{\text{old}}}(O|q)}
\frac{1}{G}\sum_{i=1}^{G}\frac{1}{|o_i|}\sum_{t=1}^{|o_i|}
\min\Big[
\rho_{i,t}A_i,\;
\mathrm{clip}\big(\rho_{i,t}, 1-\epsilon_{\mathcal{B}(i,t)}, 1+\epsilon_{\mathcal{B}(i,t)}\big)A_i
\Big],
\end{aligned}
\label{eq:ACPO}
\end{equation}

See Appendix~\ref{app:acpo} (Algorithm~\ref{app:acpo_alg}) for details.

\captionsetup[subtable]{skip=0pt} 

\begin{table*}[t]
\centering
\small
\caption{Countdown/HiTab results. Best validation reward, mean@4.
}
\label{tab:countdown_hitab}
\begin{subtable}[t]{0.45\linewidth}
\caption{Model trained on the Countdown dataset. }
\centering
\begin{tabular}{lcc}
\toprule
\multirow{2}{*}{\textbf{Method}} & \multicolumn{2}{c}{\textbf{Qwen2.5-3B-Instruct}} \\
\cmidrule(lr){2-3}
& \textbf{N-OnP.} & \textbf{OffP.} \\
\midrule
Base Model & \multicolumn{2}{c}{6.90} \\
AR-Lopti & 65.82 & 57.41\\
High-Entropy & 70.58 & 73.76 \\
Low-Entropy & 61.81	& 74.04 \\
DAPO & 73.27 & 74.38 \\
CISPO & 74.25 & 55.12 \\
ACPO & \textbf{75.74} & \textbf{76.27} \\
\bottomrule
\end{tabular}

\label{tab:countdown}
\end{subtable}
\hspace{0.03\linewidth}
\begin{subtable}[t]{0.50\linewidth}
\centering
\caption{Model trained on the HiTab dataset.}
\begin{tabular}{lcccc}
\toprule
\multirow{2}{*}{\textbf{Method}} & \multicolumn{2}{c}{\textbf{Qwen2.5-3B}} & \multicolumn{2}{c}{\textbf{Qwen2.5-7B}} \\
\cmidrule(lr){2-3} \cmidrule(lr){4-5}
& \textbf{N-OnP.} & \textbf{OffP.} & \textbf{N-OnP.} & \textbf{OffP.} \\
\midrule
Base Model & \multicolumn{2}{c}{17.50} & \multicolumn{2}{c}{31.25} \\
AR-Lopti & 45.50 &  43.33 &  68.92 & 66.58 \\
High-Entropy & 45.00 & 43.50 & 64.33 & 61.83\\
Low-Entropy & 43.17 & 43.00 & 67.00 & 65.08 \\
DAPO & 44.33 & 44.00 & 68.42 & 65.75 \\
CISPO & 43.92 & 44.83 & 69.17 & 65.67 \\
ACPO  & \textbf{45.50} & \textbf{46.42} & \textbf{69.83} & \textbf{66.58} \\
\bottomrule
\end{tabular}

\label{tab:hitab}
\end{subtable}
\end{table*}

\begin{table*}[t]
\centering
\setlength{\tabcolsep}{2pt}
\caption{Qwen2.5-7B trained on the ORZ-57K. 
AIME24/25: avg@32; others: avg@8.}
\label{tab:orz}
\resizebox{\textwidth}{!}{%
    \begin{tabular}{l cc cc cc cc cc cc cc}
    \toprule
    \multirow{2}{*}{Method} 
    & \multicolumn{2}{c}{\textbf{Minerva}} 
    & \multicolumn{2}{c}{\textbf{Math500}} 
    & \multicolumn{2}{c}{\textbf{AMC}} 
    & \multicolumn{2}{c}{\textbf{AIME24}} 
    & \multicolumn{2}{c}{\textbf{AIME25}} 
    & \multicolumn{2}{c}{\textbf{Olympiad}} 
    & \multicolumn{2}{c}{\textbf{Avg.}} \\
    \cmidrule(lr){2-3} \cmidrule(lr){4-5} \cmidrule(lr){6-7} \cmidrule(lr){8-9} \cmidrule(lr){10-11} \cmidrule(lr){12-13} \cmidrule(lr){14-15}
    & \textbf{N-OnP.} & \textbf{OffP.}
    & \textbf{N-OnP.} & \textbf{OffP.}
    & \textbf{N-OnP.} & \textbf{OffP.}
    & \textbf{N-OnP.} & \textbf{OffP.}
    & \textbf{N-OnP.} & \textbf{OffP.}
    & \textbf{N-OnP.} & \textbf{OffP.}
    & \textbf{N-OnP.} & \textbf{OffP.} \\
    \midrule
    Base Model     & \multicolumn{2}{c}{13.32} & \multicolumn{2}{c}{41.48} & \multicolumn{2}{c}{26.20} & \multicolumn{2}{c}{0.83} & \multicolumn{2}{c}{0.62} & \multicolumn{2}{c}{18.02} & \multicolumn{2}{c}{16.75} \\
    AR-Lopti       & 29.14 & 25.14 & 73.75 & 63.92 & 44.88 & 34.94 & 15.21 & 8.75  & 3.96  & 2.08  & 33.60 & 24.00 & 33.42 & 26.47 \\
    High-Entropy   & 29.08 & 29.27 & 76.88 & 75.22 & 50.75 & 46.39 & 14.37 & 14.06 & 7.81  & \textbf{10.62} & 33.54     & 34.54     & 35.41     & 35.02     \\
    Low-Entropy    & 28.72 & 25.05 & 75.85 & 72.47 & 48.95 & 40.81 & 16.04 & 11.15 & 9.48  & 4.58  & 34.69     & 26.61     & 35.62     & 30.11     \\
    DAPO           & 29.92 & 30.47 & 78.05 & 76.72 & 51.20 & 46.84 & \textbf{19.69} & \textbf{18.02} & 11.88 & 6.56  & 41.44 & 37.20 & 38.70 & 35.97 \\
    CISPO          & 30.51 & 29.55 & 78.57 & 76.30 & 49.25 & 46.08 & 18.65 & 13.44 & 10.00 & 8.33  & 40.61 & 39.10 & 37.93 & 35.47 \\
    ACPO (Ours)    & \textbf{31.02} & \textbf{31.66} & \textbf{79.78} & \textbf{80.48} & \textbf{54.37} & \textbf{51.20} & 16.77 & 16.35 & \textbf{14.37} & 10.10 & \textbf{42.83} & \textbf{42.17} & \textbf{39.86} & \textbf{38.66} \\
    \bottomrule
    \end{tabular}%
}
\end{table*}

\section{Experiments}

\subsection{Experimental Setup}
\label{sec:experimental_setup}

\textbf{Tasks, models, and training regimes.} We evaluate ACPO on three reasoning tasks spanning different domains: ORZ-57K~\citep{hu2025openreasonerzeroopensourceapproach} for mathematical problem solving, HiTab~\citep{cheng-etal-2022-hitab} for tabular question answering, and Countdown~\citep{jackson2025countdown} for arithmetic-based puzzles. Experiments use Qwen2.5-3B, Qwen2.5-3B-Instruct, and Qwen2.5-7B~\citep{qwen2.5}. Following~\citet{yang2025not,wang2025beyond}, each method is run under a near on-policy regime (N-OnP., 2 updates/rollout) and an off-policy regime (OffP., 16 updates/rollout) to test robustness across policy-divergence levels.

\textbf{Baselines, evaluation, and hyperparameters.} We compare against: (1)~\textbf{DAPO}~\citep{yu2025dapo}, asymmetric clipping with $\epsilon_h>\epsilon_l$; (2)~\textbf{CISPO}~\citep{chen2025minimax}, which clips only IS ratios rather than gating token updates; (3)~\textbf{High/Low-Entropy}~\citep{wang2025beyond}, updating only top-20\%/bottom-80\% entropy tokens; and (4)~\textbf{AR-Lopti}~\citep{yang2025not}, which scales token-wise contributions to reduce low-probability dominance. For ORZ models we evaluate on six math benchmarks (Minerva~\citep{dyer2022minerva}, Math500~\citep{hendrycks2021measuring}, AMC2023, AIME24/25, OlympiadBench~\citep{he2024olympiadbenchchallengingbenchmarkpromoting}), reporting avg@32 for AIME and avg@8 for the rest; HiTab and Countdown use avg@4 test accuracy. Statistical significance uses one-tailed Welch's $t$-test ($\alpha=0.05$). ACPO uses $B=5$ probability bins, $\alpha=3$, and $\epsilon_{\text{base}}=0.2$, with all methods sharing the same base configuration; full hyperparameters and per-method configs are in Appendix~\ref{app:detail_train}, sensitivity in Appendix~\ref{app:param_sensitivity}, and full $p$-values in Appendix~\ref{app:significance_tests}.

\subsection{Main Results}
\label{sec:main_results}
\textbf{Consistent improvements across settings.}
Tables~\ref{tab:countdown_hitab}--\ref{tab:orz} present results across three reasoning tasks (ORZ, HiTab, Countdown), multiple model scales (3B, 7B), and two training regimes (near on-policy and off-policy). ACPO achieves the best average rank of 1.22 across all 18 task-regime configurations, with 15/18 first-place and 2/18 second-place finishes, compared to DAPO (avg rank 2.56, 2/18 wins) and CISPO (avg rank 3.11, 0/18 wins). 
One-tailed Welch's $t$-tests show that ACPO's gains are statistically significant ($p<0.05$) on most benchmarks (Appendix~\ref{app:significance_tests}).

\begin{figure}[t]
\centering
\includegraphics[width=0.7\columnwidth]{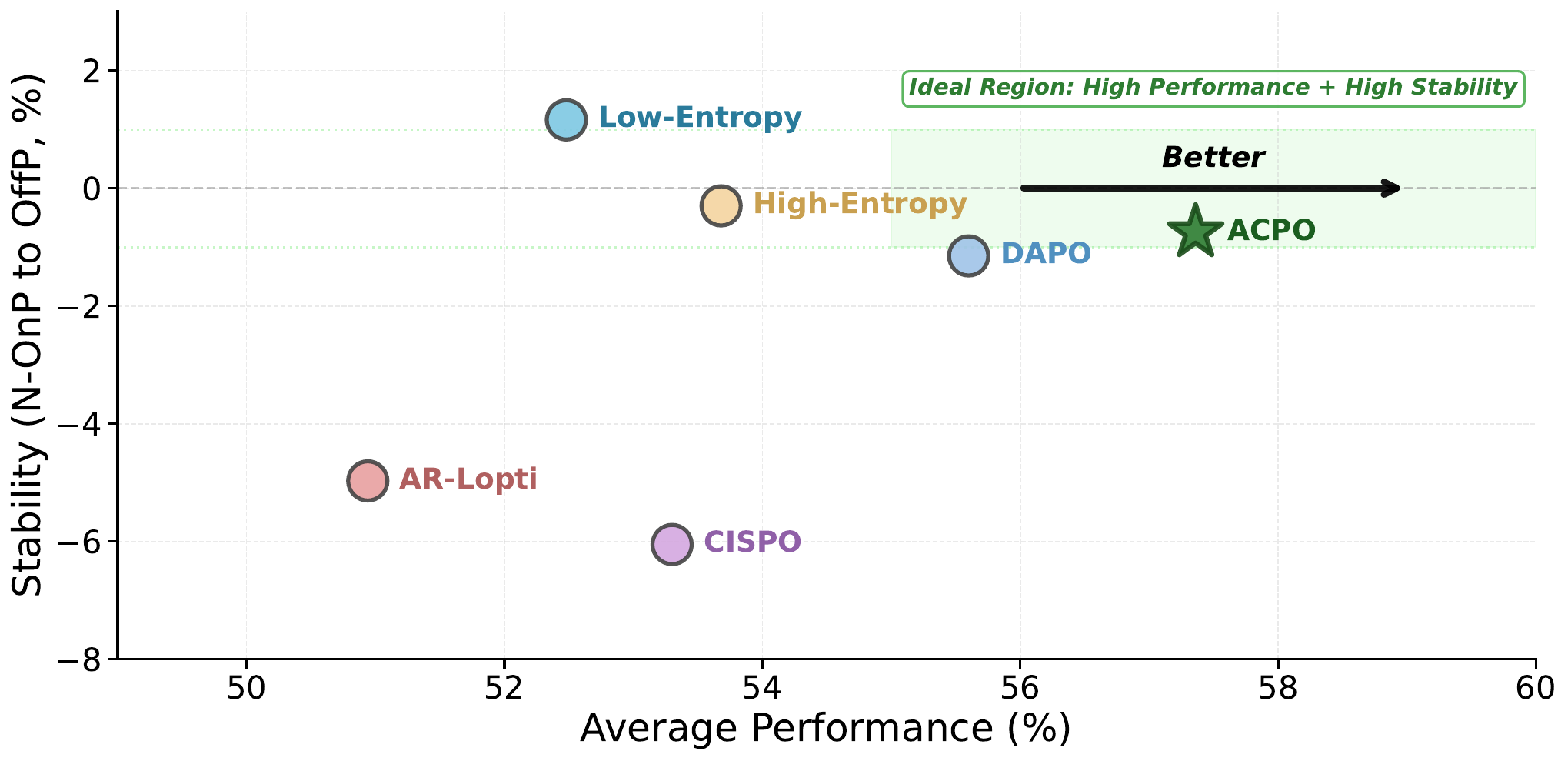}
\caption{Performance vs.\ cross-regime stability.} 
\label{fig:perf_stability}
\vskip -0.2in
\end{figure}

\textbf{Cross-regime stability.}
ACPO remains stable when moving from near on-policy to off-policy, with only a 0.8\% performance drop (Figure~\ref{fig:perf_stability}); among baselines, CISPO and AR-Lopti suffer substantial degradation ($-$6.0\% and $-$5.0\%), DAPO shows a moderate drop ($-$1.2\%), and High/Low-Entropy appear stable but at significantly lower overall performance. CISPO's instability stems from clipping IS-ratio values without gating gradient updates, while AR-Lopti's degradation reflects that suppressing low-probability tokens becomes counterproductive once gradient dominance reverses (Corollary~\ref{cor:gradient_dominance_reversal}); these results validate that ACPO's adaptive calibration to local IS-ratio dispersion supports the claim that effective RLVR requires adapting \emph{which tokens survive clipping} as off-policy degree changes rather than relying on fixed heuristics.

\textbf{Analysis of clipping and gradient dynamics.}
\label{sec:analysis}
We further analyze ACPO's clipping behavior and gradient dynamics across probability bins (Appendix~\ref{app:clip_dynamics}, Figure~\ref{fig:training_dynamics}): ACPO yields more uniform clip rates and substantially more stable gradient variance than DAPO, explaining its cross-regime stability. Appendices~\ref{sec:ablation_clip_threshold}--\ref{sec:appendix-overhead} further confirm that gains stem from adaptive calibration rather than simply using different threshold values, with negligible computational overhead.

\section{Related Works}
\textbf{Reinforcement Learning for Large Language Models.} 
RL for LLMs has shifted from preference alignment via RLHF~\citep{ouyang2022training} to directly optimizing reasoning abilities with Reinforcement Learning from Verifiable Rewards (RLVR)~\citep{lambert2024tulu3}. This paradigm uses objective feedback signals, such as mathematical correctness, to train models. Pioneered by early work like OpenAI's o1~\citep{jaech2024openai} and advanced by algorithms such as GRPO~\citep{shao2024deepseekmath}, RLVR has become a central methodology for developing state-of-the-art reasoning agents, demonstrating that complex skills can emerge from outcome-based optimization.

\textbf{Token-Level Update Strategies.} 
RLVR's success depends heavily on the token-level update mechanism. Diverse heuristics have been proposed, including entropy-based modulation~\citep{wang2025beyond, cui2025entropy}, gradient clipping~\citep{chen2025minimax, zheng2025group}, and advantage design, yet often yield conflicting conclusions. For instance, a key debate is whether to prioritize high-entropy tokens for exploration~\citep{wang2025beyond} or suppress them to reduce gradient variance~\citep{yang2025not}. Such contradictions reveal the lack of a unified understanding of RLVR update dynamics, motivating our systematic component-wise analysis.

\section{Conclusion}

We present a token-level analysis of clipped RLVR updates and identify off-policy degree as the key factor that reconciles prior conflicting heuristics.
As off-policy increases, IS-ratio dispersion and clipping reshuffle which tokens dominate the expected effective gradient, revealing the mismatch of uniform clipping.
This motivates ACPO, which uses variance-aware adaptive clipping thresholds.

Experiments across model scales (3B/7B), diverse reasoning tasks, and training regimes demonstrate that ACPO achieves consistent improvements and robust cross-regime stability, supporting our central thesis that understanding token-level gradient dynamics and adapting which tokens survive clipping accordingly, rather than simply relying on fixed heuristics, is key to effective RL for LLM reasoning.

\bibliography{neurips_2026}
\bibliographystyle{plainnat}


\appendix







\section{Reproducibility}
To ensure the reproducibility of our findings, we provide comprehensive details of our training process, including all hyperparameters and experimental configurations, in Appendix~\ref{app:detail_train}. 

\section{Ethics Statement}
The primary objective of this work is to enhance the systematic understanding and reliability of reasoning in large language models. We view this as a crucial contribution to the development of safer and more predictable AI. All experiments were conducted on publicly available, standard academic datasets and utilized open-source base models. These resources do not contain any personally identifiable information or sensitive user data. While the base models may inherit societal biases from their pre-training data, our research focuses on optimizing for objective, verifiable tasks such as mathematical and logical correctness. This problem formulation is less susceptible to subjective biases, though we acknowledge that addressing underlying model bias remains a critical and ongoing challenge for the community. We also recognize that advancements in AI capabilities carry a potential for dual use, and we encourage the responsible development and deployment of these technologies. Our work is intended for a research audience to foster a deeper, more principled understanding of AI systems.

\section{Detailed Theoretical Derivations}
\subsection{Derivation of Lemma \ref{lemma:is_ratio_variance}}
\label{app:derivation_lemma}
This appendix derives the relationship between the variance of the importance sampling ratio, $\sigma_\rho^2$, and the token probability $\pi_k$ in Lemma~\ref{lemma:is_ratio_variance}.

\subsubsection{A small-step approximation in logit space}
Let $\pi_k$ be produced by a softmax with temperature $T$ over logits $y\in\mathbb{R}^V$:
$\pi_k = \frac{\exp(y_k/T)}{\sum_j \exp(y_j/T)}$.
A single update step perturbs the entire logit vector by $\Delta y\in\mathbb{R}^V$, so $\Delta\pi_k$ is driven by changes to \emph{all} logits, not just $y_k$.

The relevant softmax partial derivatives are
\begin{equation}
    \frac{\partial \pi_k}{\partial y_k} = \frac{\pi_k(1-\pi_k)}{T},
    \qquad
    \frac{\partial \pi_k}{\partial y_j} = -\frac{\pi_k\,\pi_j}{T}\quad (j\ne k).
\end{equation}
Combining the contributions of all logits, the first-order expansion is
\begin{equation}
\Delta\pi_k \;\approx\; \frac{\pi_k}{T}\Big[(1-\pi_k)\,\Delta y_k - \sum_{j\ne k}\pi_j\,\Delta y_j\Big].
\label{eq:full_taylor}
\end{equation}
The IS ratio for token $k$ is $\rho_k=\pi'_k/\pi_k=1+\Delta\pi_k/\pi_k$, so
\begin{equation}
\rho_k \;\approx\; 1 + \frac{1}{T}\Big[(1-\pi_k)\,\Delta y_k - \sum_{j\ne k}\pi_j\,\Delta y_j\Big].
\label{eq:rho_linearized}
\end{equation}

\subsubsection{Variance of the Importance Sampling Ratio}
We now compute the variance of $\rho_k$ from Eq.~\eqref{eq:rho_linearized}. The IS ratio $\rho_k$ is a random variable whose randomness comes from two sources: (i) the per-step logit perturbation $\Delta y$, and (ii) the policy state itself, i.e., the full vocabulary distribution $\boldsymbol\pi=(\pi_k,\boldsymbol\pi_{\ne k})$ that varies across prompts and positions. We will first take variance with respect to $\Delta y$ at a fixed $\boldsymbol\pi$, then marginalize over $\boldsymbol\pi_{\ne k}$ to obtain a single scalar function of $\pi$.

\paragraph{Assumption (per-step logit perturbation).}
We treat the per-step logit perturbation $\Delta y$ as a random vector with zero mean, common per-coordinate variance $\sigma_y^2$, and uncorrelated coordinates:
\begin{equation}
    \mathbb{E}[\Delta y_j]=0,\quad
    \mathrm{Var}(\Delta y_j) = \sigma_y^2, \quad
    \mathrm{Cov}(\Delta y_j,\Delta y_{j'})=0\ \ (j\ne j').
\end{equation}
The scalar $\sigma_y^2$ summarizes the magnitude of one update step in logit space and serves as a single per-step drift scale.

\paragraph{Step 1: conditional variance given the policy state.}
Conditioning on the policy state $\boldsymbol\pi$, the variance of Eq.~\eqref{eq:rho_linearized} over $\Delta y$ is
\begin{align}
    \sigma_\rho^2(\boldsymbol\pi)
    &\triangleq \mathrm{Var}(\rho_k \mid \boldsymbol\pi)
    = \frac{1}{T^2}\Big[\,(1-\pi_k)^2\,\mathrm{Var}(\Delta y_k) + \sum_{j\ne k}\pi_j^2\,\mathrm{Var}(\Delta y_j)\Big]
    + R \\
    &= \frac{\sigma_y^2}{T^2}\Big[\,(1-\pi_k)^2 + \sum_{j\ne k}\pi_j^2\,\Big] + R,
\label{eq:full_var}
\end{align}
where $R$ is the contribution of higher-order Taylor terms in $\Delta y$ (discussed below).

Define the \emph{shape factor} of the residual distribution over non-sampled tokens by
\begin{equation}
    S(\boldsymbol{\pi}_{\ne k}) \;\triangleq\; \sum_{j\ne k}q_j^2,
    \qquad q_j \;\triangleq\; \frac{\pi_j}{1-\pi_k}\quad(j\ne k),
\end{equation}
so $\{q_j\}_{j\ne k}$ is a normalized distribution and $S\in[\,1/(V-1),\,1\,]$. Substituting $\sum_{j\ne k}\pi_j^2=(1-\pi_k)^2\,S$ gives the factorized form
\begin{equation}
    \sigma_\rho^2(\boldsymbol\pi) \;=\; \frac{\sigma_y^2}{T^2}\,(1-\pi_k)^2\,\bigl[1+S(\boldsymbol{\pi}_{\ne k})\bigr] + R,
\label{eq:var_factorized}
\end{equation}
which separates a clean $\pi_k$-scaling, $(1-\pi_k)^2$, from a bounded distribution-shape factor $1+S\in[1,2]$.

\paragraph{Step 2: order of the remainder $R$.}
Writing the full Taylor expansion of $\rho_k$ in $\Delta y$ as $\rho_k = 1 + L(\Delta y) + Q(\Delta y) + \cdots$ with $L$ linear and $Q$ quadratic, we have
\begin{equation}
    \mathrm{Var}(\rho_k\mid\boldsymbol\pi) = \mathrm{Var}(L) + 2\,\mathrm{Cov}(L,Q) + \mathrm{Var}(Q) + \cdots
\end{equation}
The leading term $\mathrm{Var}(L)$ is $O(\sigma_y^2)$ and gives Eq.~\eqref{eq:var_factorized}. The cross term $\mathrm{Cov}(L,Q)$ is controlled by the third moment of $\Delta y$ and is $O(\sigma_y^3)$; $\mathrm{Var}(Q)$ is $O(\sigma_y^4)$. Hence $R=O(\sigma_y^3)$ in general, and degenerates to $O(\sigma_y^4)$ when $\Delta y$ is symmetric.

\paragraph{Step 3: marginalize over $\boldsymbol\pi_{\ne k}$ to obtain $\sigma_\rho^2(\pi)$.}
The shape factor $S(\boldsymbol\pi_{\ne k})$ depends on the full vocabulary distribution, not just on $\pi_k$. To obtain the single-variable summary $\sigma_\rho^2(\pi)$ used in the main text, we average Eq.~\eqref{eq:var_factorized} over the conditional distribution of $\boldsymbol\pi_{\ne k}$ given $\pi_k=\pi$:
\begin{equation}
    \sigma_\rho^2(\pi) \;\triangleq\; \mathbb{E}\bigl[\,\sigma_\rho^2(\boldsymbol\pi)\,\big|\,\pi_k=\pi\,\bigr]
    \;=\; \frac{\sigma_y^2}{T^2}\,(1-\pi)^2\,\bigl(1+\bar{S}(\pi)\bigr) + \mathbb{E}[R\mid\pi_k=\pi],
\end{equation}
where $\bar S(\pi)\triangleq\mathbb{E}[S(\boldsymbol\pi_{\ne k})\mid\pi_k=\pi]\in[\,1/(V-1),\,1\,]$.

Because $1+\bar S(\pi)$ is a bounded factor in $[1,2]$ and varies mildly compared with $(1-\pi)^2$, we absorb its (overall) magnitude into an \emph{effective} per-step logit-perturbation scale,
\begin{equation}
    \kappa^2 \;\triangleq\; \frac{\sigma_y^2}{T^2}\bigl(1+\bar{S}\bigr),
    \qquad \bar S\triangleq\mathbb{E}_\pi[\bar S(\pi)],
\end{equation}
which yields an exact two-sided bound up to a factor of two,
\begin{equation}
    \tfrac{1}{2}\,\kappa^2(1-\pi)^2 \;\le\; \frac{\sigma_y^2}{T^2}(1-\pi)^2 \;\le\; \sigma_\rho^2(\pi) - \mathbb{E}[R\mid\pi_k=\pi] \;\le\; \frac{2\sigma_y^2}{T^2}(1-\pi)^2 \;\le\; 2\kappa^2(1-\pi)^2.
\end{equation}
With this convention,
\begin{equation}
    \sigma_\rho^2(\pi) \;=\; \kappa^2(1-\pi)^2 \;+\; O(\kappa^3),
\label{eq:var_final}
\end{equation}
which matches the form stated in Lemma~\ref{lemma:is_ratio_variance}. The $O(\kappa^3)$ remainder collects (i) the higher-order Taylor remainder $\mathbb{E}[R\mid\pi_k=\pi]$ from Step 2, and (ii) the residual variation of $1+\bar S(\pi)$ around its average $1+\bar S$. Both are negligible under the standard small-step assumption.

The leading term shows that $\sigma_\rho^2$ decreases (approximately) monotonically with $\pi$ on $(0,1)$: low-probability tokens have substantially wider IS-ratio distributions and are thus more susceptible to clipping.

\subsubsection{Empirical validation of the variance scaling}
\begin{figure}[t]
    \centering
    \includegraphics[width=0.50\textwidth]{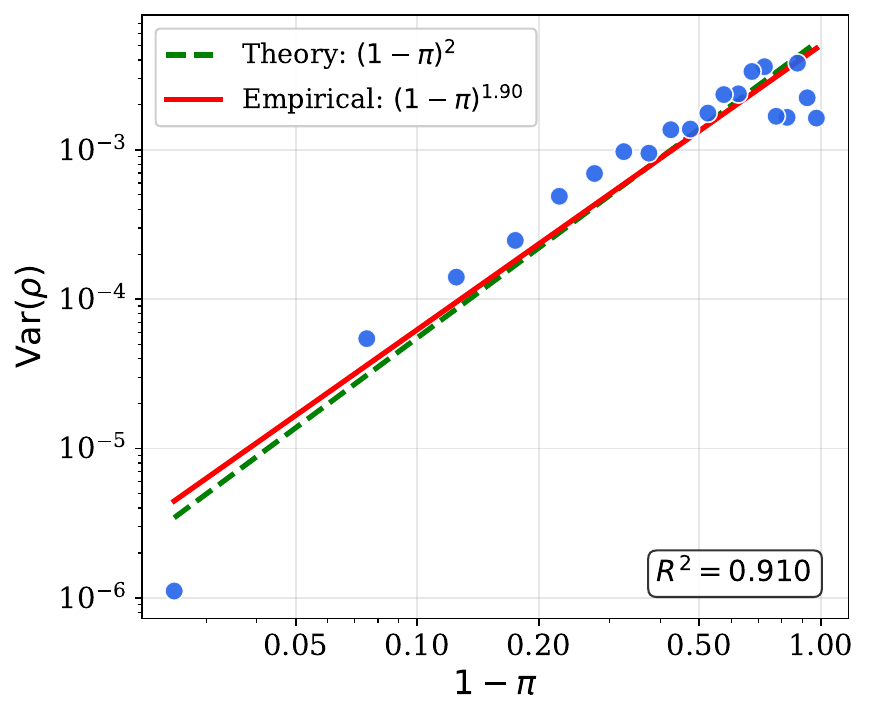}
    \caption{Empirical validation of Lemma~\ref{lemma:is_ratio_variance} via log-log regression. Green: theoretical scaling $\mathrm{Var}(\rho)\propto(1-\pi)^2$. Red: fitted curve $\mathrm{Var}(\rho)=5.01\times10^{-3}(1-\pi)^{1.90}$ (slope $1.90\pm0.14$, $R^2=0.91$).}
    \label{fig:var_vs_prob_single}
\end{figure}

To validate the prediction $\mathrm{Var}(\rho)\propto (1-\pi)^2$, we perform a log-log regression on binned tokens.
As shown in Figure~\ref{fig:var_vs_prob_single}, the fitted exponent is $1.90\pm 0.14$ with $R^2=0.91$ (20 bins; 54,173 tokens), which closely matches the theoretical exponent of 2.
The green curve indicates the theoretical scaling, while the red curve shows the empirical fit:
\begin{equation}
\mathrm{Var}(\rho) = 5.01 \times 10^{-3}\cdot (1-\pi)^{1.90}.
\end{equation}

\subsection{Derivation of Proposition \ref{prop:gradient_expectation}}
\label{app:derivation_proposition}

This appendix derives the closed form for $\mathbb{E}[G\mid\pi]$ stated in Proposition~\ref{prop:gradient_expectation}.
We work under the three modeling assumptions stated in Section~\ref{sec:explain_the_difference}, restated here for convenience:
\begin{itemize}[leftmargin=*]
    \item[(M1)] \emph{Log-normal IS-ratio.} $\log\rho\mid\pi\sim\mathcal{N}\!\big(-\tau^2(\pi)/2,\,\tau^2(\pi)\big)$, with $\tau^2(\pi)=\log\!\big(1+\sigma_\rho^2(\pi)\big)$. The log-normal choice is for closed-form convenience; Appendix~\ref{app:gaussian_robustness} reproduces the qualitative conclusions (small-$\bar\sigma_\rho^2$ scaling, dominance reversal) under a Gaussian IS-ratio model.
    \item[(M2)] \emph{Linear-coupling ansatz.} $A=\mu(\pi)\,u+\eta$, where $u:=\log\rho+\tau^2(\pi)/2$ is the centered log-ratio, $\mu(\pi)\ge 0$, and $\mathbb{E}[\eta\mid u,\pi]=0$.
    \item[(M3)] \emph{Perfect-coupling idealization.} The closed form below is computed under $\eta\equiv 0$ (the finite-$\eta$ effect is discussed at the end of this section).
\end{itemize}

\subsubsection{Setup and the trust-region indicator}
The token-level effective coefficient (Eq.~\ref{eq:sim_grad}) is
\begin{equation}
G \;=\; (1-\pi)\,\rho\,A\,\mathbb{I}_{\text{trust}}(\rho,A),
\qquad
\mathbb{I}_{\text{trust}}(\rho,A) =
\begin{cases}
1, & A>0,\ \rho\le 1+\epsilon_h,\\
1, & A<0,\ \rho\ge 1-\epsilon_l,\\
0, & \text{otherwise.}
\end{cases}
\label{eq:app_indicator_trust}
\end{equation}
Note that the indicator depends on $A$ through $\mathrm{sign}(A)$: positive advantages are clipped from above, negative advantages from below. Under (M3), $\eta\equiv 0$ implies $A=\mu(\pi)\,u$ with $\mu(\pi)\ge 0$, so $\mathrm{sign}(A)=\mathrm{sign}(u)$ wherever $\mu(\pi)>0$.\footnote{If $\mu(\pi)=0$ on a measure-zero set, $G$ vanishes there and the closed form is unaffected.}
Therefore, the indicator collapses to a condition on the centered log-ratio $u$ alone:
\begin{equation}
\mathbb{I}_{\text{trust}}\Big|_{\eta\equiv 0}
\;=\; \mathbf{1}\{u>0,\ \log\rho\le \log(1+\epsilon_h)\}
\;+\; \mathbf{1}\{u<0,\ \log\rho\ge \log(1-\epsilon_l)\}.
\label{eq:app_indicator_perfect}
\end{equation}

\subsubsection{Standardization}
Under (M1), the centered log-ratio $u=\log\rho+\tau^2/2$ satisfies $u\mid\pi\sim\mathcal{N}(0,\tau^2)$. Define the standardized variable
\begin{equation}
Z \;:=\; u/\tau \;=\; \big(\log\rho+\tau^2/2\big)/\tau \;\sim\; \mathcal{N}(0,1).
\end{equation}
The two log-ratio thresholds $\log(1\pm\epsilon)$ map to standardized endpoints
\begin{equation}
L(\pi) \;=\; \frac{\log(1-\epsilon_l)+\tau^2/2}{\tau},
\qquad
U(\pi) \;=\; \frac{\log(1+\epsilon_h)+\tau^2/2}{\tau}.
\label{eq:app_LU}
\end{equation}
For typical clip parameters and small $\tau$, $L(\pi)<0<U(\pi)$. We do not assume this in general; we use $L'(\pi):=\min(L(\pi),0)$ to keep the lower piece of the integration domain inside $\{u<0\}$.

Substituting $\rho=\exp(\tau Z-\tau^2/2)$ and $A=\mu(\pi)\,\tau Z$ into Eq.~\eqref{eq:app_indicator_perfect}, the integration region in $Z$ becomes
\begin{equation}
\Omega(\pi) \;=\; \{Z:0<Z\le U(\pi)\}\ \cup\ \{Z:L'(\pi)\le Z<0\}.
\label{eq:app_omega}
\end{equation}
(When $L\ge 0$, the lower piece $\{L\le Z<0\}$ is empty; using $L'=\min(L,0)$ keeps the closed form uniform across the regimes $L<0$ and $L\ge 0$.)

\subsubsection{Computing $\mathbb{E}[G\mid\pi]$}
Under (M3), Eq.~\eqref{eq:app_indicator_perfect} and the substitutions above give
\begin{align}
\mathbb{E}[G\mid\pi]
&= (1-\pi)\,\mathbb{E}\!\left[\rho\cdot A\cdot \mathbb{I}_{\text{trust}}\,\Big|\,\pi\right] \notag\\
&= (1-\pi)\,\mu(\pi)\,
\mathbb{E}\!\left[\,e^{\tau Z-\tau^2/2}\cdot \tau Z\cdot \mathbf{1}_{Z\in\Omega(\pi)}\,\right] \notag\\
&= (1-\pi)\,\mu(\pi)\,\tau\;\underbrace{\int_{\Omega(\pi)} z\,e^{\tau z-\tau^2/2}\,\phi(z)\,dz}_{=:\,J(\tau,\pi)}.
\label{eq:app_J_def}
\end{align}
We compute $J(\tau,\pi)$ using the completion-of-square identity for the standard normal density:
\begin{equation}
e^{\tau z}\,\phi(z) \;=\; \frac{1}{\sqrt{2\pi}}\exp\!\Big(\tau z-\tfrac{z^2}{2}\Big)
\;=\; \frac{1}{\sqrt{2\pi}}\exp\!\Big(\tfrac{\tau^2}{2}-\tfrac{(z-\tau)^2}{2}\Big)
\;=\; e^{\tau^2/2}\,\phi(z-\tau).
\label{eq:app_completion}
\end{equation}
Therefore $e^{\tau z-\tau^2/2}\phi(z)=\phi(z-\tau)$, and
\begin{equation}
J(\tau,\pi) \;=\; \int_{\Omega(\pi)} z\,\phi(z-\tau)\,dz.
\label{eq:app_J_simplified}
\end{equation}
Using the elementary identity $\int z\,\phi(z-\tau)\,dz = -\phi(z-\tau)+\tau\,\Phi(z-\tau)+C$, we evaluate the two pieces of $\Omega(\pi)=[L'(\pi),0)\cup(0,U(\pi)]$:
\begin{align}
\int_{L'(\pi)}^{0} z\,\phi(z-\tau)\,dz
&= \big[-\phi(z-\tau)+\tau\Phi(z-\tau)\big]_{L'(\pi)}^{0} \notag\\
&= -\phi(-\tau)+\tau\Phi(-\tau)
\;+\;\phi(L'-\tau)-\tau\Phi(L'-\tau), \\
\int_{0}^{U(\pi)} z\,\phi(z-\tau)\,dz
&= -\phi(U-\tau)+\tau\Phi(U-\tau)
\;+\;\phi(-\tau)-\tau\Phi(-\tau).
\end{align}
Adding the two pieces, the boundary contributions at $z=0$ cancel exactly, yielding
\begin{equation}
J(\tau,\pi) \;=\; \phi(L'(\pi)-\tau)-\phi(U(\pi)-\tau)\;+\;\tau\big[\Phi(U(\pi)-\tau)-\Phi(L'(\pi)-\tau)\big].
\label{eq:app_J_closed}
\end{equation}

\subsubsection{The closed form $H(\tau)$}
Combining Eq.~\eqref{eq:app_J_def} and Eq.~\eqref{eq:app_J_closed}, we obtain
\begin{equation}
\mathbb{E}[G\mid\pi]
\;=\; (1-\pi)\,\mu(\pi)\,H\!\big(\tau(\pi)\big),
\quad
H(\tau)
\;=\; \tau\Big\{\phi(L'-\tau)-\phi(U-\tau)+\tau\big[\Phi(U-\tau)-\Phi(L'-\tau)\big]\Big\},
\label{eq:app_closed_form}
\end{equation}
which matches Proposition~\ref{prop:gradient_expectation}. The three factors admit the following interpretation: $(1-\pi)$ is the on-policy logit-gradient front factor; $\mu(\pi)$ is the local advantage--log-ratio coupling under (M2); $H(\tau)$ aggregates the trust-region indicator's effect on the tilted normal density (whose mean is shifted from $0$ to $\tau$ by the IS factor) over the standardized window $[L',U]$.

\subsubsection{Small-$\tau$ Taylor expansion}
\label{app:H_taylor}
For small $\tau$, the standardized endpoints behave as
\begin{equation}
U(\pi) \;=\; \frac{\log(1+\epsilon_h)}{\tau}+\frac{\tau}{2} \;\to\; +\infty,
\qquad
L(\pi) \;=\; \frac{\log(1-\epsilon_l)}{\tau}+\frac{\tau}{2} \;\to\; -\infty,
\end{equation}
so $\Phi(U-\tau)\to 1$, $\Phi(L'-\tau)\to 0$, and $\phi(U-\tau),\phi(L'-\tau)\to 0$ super-polynomially. Hence
\begin{equation}
H(\tau) \;=\; \tau\cdot \tau\cdot\big(1-0\big)+o(\tau^2) \;=\; \tau^2+O(\tau^4).
\label{eq:app_H_taylor}
\end{equation}
Substituting Lemma~\ref{lemma:is_ratio_variance}, $\sigma_\rho^2(\pi)=\kappa^2(1-\pi)^2+O(\kappa^3)$, so $\tau^2(\pi)=\log(1+\sigma_\rho^2(\pi))=\sigma_\rho^2(\pi)+O\!\big(\sigma_\rho^4(\pi)\big)$. This gives the near-on-policy asymptotic
\begin{equation}
\mathbb{E}[G\mid\pi] \;=\; \mu(\pi)(1-\pi)\,\sigma_\rho^2(\pi) \;+\; O\!\big(\sigma_\rho^4(\pi)\big)
\;=\; \mu(\pi)(1-\pi)^3\,\kappa^2 \;+\; O(\kappa^4),
\label{eq:app_small_kappa}
\end{equation}
where the second equality substitutes Lemma~\ref{lemma:is_ratio_variance} for explicit $\pi$-scaling.
This recovers a clean ``$(1-\pi)^3$ scaling'' regime used in §\ref{app:derivation_corollary} to characterize the small-$\bar\sigma_\rho^2$ behavior of $D$.

\subsubsection{Large-$\tau$ behavior}
For large $\tau$, the endpoints satisfy
\begin{equation}
U(\pi)-\tau \;=\; \frac{\log(1+\epsilon_h)}{\tau}-\frac{\tau}{2}\;\to\;-\infty,
\qquad
L'(\pi)-\tau \;\le\; -\tau\;\to\;-\infty,
\end{equation}
so $\Phi(U-\tau)-\Phi(L'-\tau)\to 0$ and $H(\tau)\to 0$. Combined with the small-$\tau$ growth $H(\tau)\approx\tau^2$, $H$ is unimodal in $\tau$ for typical $\epsilon_h,\epsilon_l$, peaks at an intermediate $\tau$, and eventually decreases as $\tau\to\infty$. Translating to the local off-policy axis, $\tau^2(\pi)=\log(1+\sigma_\rho^2(\pi))$ grows only logarithmically in $\sigma_\rho^2(\pi)$, so the suppression of $H$ as a function of $\sigma_\rho^2(\pi)$ (and hence of $\bar\sigma_\rho^2$) is polynomial, not exponential.

\subsubsection{Numerical illustration setup}
\label{app:numerical_setup}
The closed form Eq.~\eqref{eq:app_closed_form} is fully specified once $(\mu(\pi),p(\pi),\epsilon_h,\epsilon_l)$ and the off-policy degree (parametrized by $\kappa$ via Lemma~\ref{lemma:is_ratio_variance}, or equivalently by $\bar\sigma_\rho^2=\kappa^2\,\mathbb{E}_{p(\pi)}[(1-\pi)^2]+O(\kappa^3)$) are chosen. For the numerical illustrations referenced in this paper, we use clip parameters $(\epsilon_h,\epsilon_l)=(0.2,0.2)$ matching the experimental defaults; representative choices for $p(\pi)$ are taken from the empirical token-probability histogram of a Qwen2.5-7B rollout on math-reasoning prompts; and $\mu(\pi)$ is treated as a slowly-varying non-negative function (we use $\mu(\pi)\equiv\mu_0$ as the simplest representative, which already exhibits the qualitative behavior described in §\ref{app:derivation_corollary}). These choices are intended for transparency, not parameter fitting; the qualitative phenomena ($H(\tau)$ unimodality, polynomial suppression in $\bar\sigma_\rho^2$, and the dominance reversal in §\ref{app:derivation_corollary}) are not sensitive to them.

\subsubsection{Finite-$\eta$ caveat}
\label{app:finite_eta}
The closed form Eq.~\eqref{eq:app_closed_form} is derived under (M3) ($\eta\equiv 0$), which makes $\mathrm{sign}(A)=\mathrm{sign}(u)$ and collapses the trust-region indicator to a region in $u$ alone. With finite $\eta$, $\mathrm{sign}(A)$ and $\mathrm{sign}(u)$ can disagree, and Eq.~\eqref{eq:app_indicator_perfect} is replaced by truncated-normal moments of $A$ given $\rho$. We do not develop the finite-$\eta$ extension here and use (M3) only as a tractable analytical guide for the qualitative phenomena above.

\subsubsection{Robustness to the IS-ratio distributional choice}
\label{app:gaussian_robustness}
The log-normal model in (M1) is a closed-form-friendly choice: it enforces $\rho>0$ and $\mathbb{E}[\rho\mid\pi]=1$ while reducing the trust-region integral to a tilted Gaussian. The qualitative conclusions used in this paper, however, do not rely on the specific distributional choice. To make this concrete, we re-derive the relevant quantities under a Gaussian IS-ratio model and verify that the same $\bar\sigma_\rho^2$-axis behavior emerges.

\paragraph{Gaussian IS-ratio model.}
Replace (M1) with
\begin{equation}
\rho\mid\pi\;\sim\;\mathcal{N}\!\big(1,\,\sigma_\rho^2(\pi)\big),
\tag{M1$'$}
\end{equation}
keeping (M2) (linear coupling) in the form $A=\mu(\pi)(\rho-1)+\eta$ and (M3) ($\eta\equiv 0$). Under (M3), $\mathrm{sign}(A)=\mathrm{sign}(\rho-1)$, so the trust-region indicator collapses to $\mathbf{1}\{\rho\in[1-\epsilon_l,\,1+\epsilon_h]\}$.

Standardize $z:=(\rho-1)/\sigma_\rho(\pi)\sim\mathcal{N}(0,1)$, with endpoints $a:=-\epsilon_l/\sigma_\rho(\pi)$, $b:=\epsilon_h/\sigma_\rho(\pi)$. Substituting and using the identities $\int_a^b z\phi(z)\,dz=\phi(a)-\phi(b)$ and $\int_a^b z^2\phi(z)\,dz=a\phi(a)-b\phi(b)+\Phi(b)-\Phi(a)$,
\begin{equation}
\mathbb{E}[G\mid\pi]
\;=\; (1-\pi)\,\mu(\pi)\,\sigma_\rho^2(\pi)\,\widetilde H\!\big(\sigma_\rho(\pi)\big),
\label{eq:app_gaussian_closed_form}
\end{equation}
where
\begin{equation}
\widetilde H(\sigma_\rho)
\;:=\;
\big[\Phi(b)-\Phi(a)\big]
\;+\;\big[a\phi(a)-b\phi(b)\big]
\;+\;\frac{1}{\sigma_\rho}\big[\phi(a)-\phi(b)\big].
\end{equation}
Equation~\eqref{eq:app_gaussian_closed_form} plays the role of Eq.~\eqref{eq:app_closed_form} for the Gaussian model.

\paragraph{Small-$\bar\sigma_\rho^2$ scaling.}
For small $\sigma_\rho(\pi)$, $a\to-\infty$ and $b\to+\infty$, so $\phi(a),\phi(b),a\phi(a),b\phi(b)\to 0$ and $\Phi(b)-\Phi(a)\to 1$. Hence $\widetilde H\!\big(\sigma_\rho(\pi)\big)\to 1$, and
\begin{equation}
\mathbb{E}[G\mid\pi] \;=\; (1-\pi)\,\mu(\pi)\,\sigma_\rho^2(\pi) + O\!\big(e^{-c/\sigma_\rho^2}\big)
\;=\; \mu(\pi)(1-\pi)^3\,\kappa^2 + O(\kappa^4),
\label{eq:app_gaussian_small}
\end{equation}
where the second equality uses Lemma~\ref{lemma:is_ratio_variance}. This matches the leading-order behavior under (M1) (Eq.~\eqref{eq:app_small_kappa}) exactly, so the (C1) limit in §\ref{app:C1_proof} is identical: $\widetilde C_0$ and $C_0$ are the same structural constants under either distributional choice.

\paragraph{Large-$\sigma_\rho$ behavior.}
For large $\sigma_\rho(\pi)$, $a,b\to 0$, so $\Phi(b)-\Phi(a)\to 0$, $a\phi(a)-b\phi(b)\to 0$, and $\phi(a)-\phi(b)=O(\epsilon_h+\epsilon_l)/\sigma_\rho$. Hence $\widetilde H\to 0$ and $\mathbb{E}[G\mid\pi]\to 0$, giving the same trust-region suppression mechanism that drives (C2). The Gaussian model has the well-known artifact that $\rho$ can take negative values when $\sigma_\rho^2(\pi)$ is comparable to $1$, which is the principal practical reason we adopt log-normal; the IVT argument and the dominance reversal, however, do not rely on positivity of $\rho$.

\paragraph{Takeaway.}
The same closed-form factorization $\mathbb{E}[G\mid\pi]=(1-\pi)\mu(\pi)f\!\big(\sigma_\rho(\pi)\big)$, the same near-on-policy $(1-\pi)^3\,\kappa^2$ scaling, the same $C_0$, and the same large-$\sigma_\rho$ suppression hold under either (M1) or (M1$'$). The dominance reversal in §\ref{app:derivation_corollary} therefore does not rely on the log-normal choice; (M1) is selected to keep the closed form clean (especially for $\rho>0$), not as a load-bearing assumption.

\subsection{Derivation of Corollary \ref{cor:gradient_dominance_reversal}}
\label{app:derivation_corollary}
This appendix derives the sufficient condition for the gradient dominance reversal stated in Corollary~\ref{cor:gradient_dominance_reversal}, using the closed form from Appendix~\ref{app:derivation_proposition}.

\subsubsection{Setup}
By Proposition~\ref{prop:gradient_expectation},
\begin{equation}
\mathbb{E}[G\mid\pi] \;=\; (1-\pi)\,\mu(\pi)\,H\!\big(\tau(\pi)\big),
\qquad
\tau^2(\pi) \;=\; \log\!\big(1+\sigma_\rho^2(\pi)\big),
\end{equation}
with $H(\tau)$ defined in Eq.~\eqref{eq:app_closed_form}. Fix $0<p_L<p_H<1$ and let $I_L=[0,p_L]$, $I_H=[p_H,1)$. Following the main text, we define the group-level conditional averages with respect to the rollout token-probability density $p(\pi)$:
\begin{align}
\bar G_L &\;=\; \mathbb{E}\!\big[\,\mathbb{E}[G\mid\pi]\,\big|\,\pi\in I_L\big]
\;=\; \frac{1}{\mathbb{P}(\pi\in I_L)}\int_{0}^{p_L}\mathbb{E}[G\mid\pi]\,p(\pi)\,d\pi, \\
\bar G_H &\;=\; \mathbb{E}\!\big[\,\mathbb{E}[G\mid\pi]\,\big|\,\pi\in I_H\big]
\;=\; \frac{1}{\mathbb{P}(\pi\in I_H)}\int_{p_H}^{1}\mathbb{E}[G\mid\pi]\,p(\pi)\,d\pi.
\end{align}
We assume $\mathbb{P}(\pi\in I_L),\mathbb{P}(\pi\in I_H)>0$ throughout.

\paragraph{Off-policy axis.}
The main-text statement of Corollary~\ref{cor:gradient_dominance_reversal} parametrizes $D$ by the global off-policy degree $\bar\sigma_\rho^2:=\mathbb{E}_{p(\pi)}[\sigma_\rho^2(\pi)]$. To carry out the analysis, we use Lemma~\ref{lemma:is_ratio_variance}, $\sigma_\rho^2(\pi)=\kappa^2(1-\pi)^2+O(\kappa^3)$, which gives
\begin{equation}
\bar\sigma_\rho^2 \;=\; \kappa^2\,c_p \;+\; O(\kappa^3),
\qquad c_p:=\mathbb{E}_{p(\pi)}\big[(1-\pi)^2\big]\;>\;0.
\label{eq:app_sigmabar_kappa}
\end{equation}
Hence $\bar\sigma_\rho^2$ is a strictly increasing function of $\kappa^2$ for small $\kappa$, and the IVT statement on $\bar\sigma_\rho^2$ is equivalent to the corresponding IVT statement on $\kappa$. In what follows we carry out the proof in the natural parametrization $\kappa$ and translate the conclusions back to $\bar\sigma_\rho^2$. Define the difference
\begin{equation}
D(\kappa) \;:=\; \bar G_L\big|_{\kappa}-\bar G_H\big|_{\kappa},
\qquad
D(\bar\sigma_\rho^2) \;=\; D(\kappa(\bar\sigma_\rho^2)),
\label{eq:app_D_def}
\end{equation}
where $\kappa(\bar\sigma_\rho^2)=\sqrt{\bar\sigma_\rho^2/c_p}+O(\bar\sigma_\rho^2)$ inverts Eq.~\eqref{eq:app_sigmabar_kappa}.

\subsubsection{(C1) Small-$\bar\sigma_\rho^2$ behavior of $D$}
\label{app:C1_proof}
By Eq.~\eqref{eq:app_small_kappa}, for each fixed $\pi\in(0,1)$,
\begin{equation}
\mathbb{E}[G\mid\pi] \;=\; \mu(\pi)(1-\pi)^3\,\kappa^2 \;+\; r(\pi,\kappa),
\qquad r(\pi,\kappa)=O(\kappa^4),
\end{equation}
uniformly on compact subsets of $(0,1)$. Integrating against $p(\pi)$ over $I_L$ and $I_H$,
\begin{equation}
\bar G_L\big|_{\kappa} \;=\; \kappa^2\,\mathbb{E}\!\left[\mu(\pi)(1-\pi)^3\,\big|\,\pi\in I_L\right] \;+\; O(\kappa^4),
\end{equation}
and similarly for $\bar G_H\big|_{\kappa}$. Subtracting,
\begin{equation}
\frac{D(\kappa)}{\kappa^2}
\;\xrightarrow[\kappa\to 0^+]{}\;
\widetilde C_{0}
\;:=\;
\mathbb{E}\!\left[\mu(\pi)(1-\pi)^3\,\big|\,\pi\in I_L\right]
\;-\;
\mathbb{E}\!\left[\mu(\pi)(1-\pi)^3\,\big|\,\pi\in I_H\right].
\label{eq:app_C1_kappa_limit}
\end{equation}
By Eq.~\eqref{eq:app_sigmabar_kappa}, $\kappa^2=\bar\sigma_\rho^2/c_p+O(\bar\sigma_\rho^4)$, so dividing by $\bar\sigma_\rho^2$ instead gives
\begin{equation}
\frac{D(\bar\sigma_\rho^2)}{\bar\sigma_\rho^2}
\;\xrightarrow[\bar\sigma_\rho^2\to 0^+]{}\;
C_0 \;:=\; \frac{\widetilde C_{0}}{c_p}.
\label{eq:app_C1_limit}
\end{equation}
\emph{(C1) is the assumption $C_{0}>0$, equivalently $\widetilde C_{0}>0$.} Since $(1-\pi)^3$ is strictly larger on $I_L$ than on $I_H$ (as $p_L<p_H$ implies $\inf_{I_L}(1-\pi)>\sup_{I_H}(1-\pi)$), $C_{0}>0$ holds whenever $\mu(\pi)$ is bounded below by a positive constant on $I_L$. More generally, (C1) holds whenever $\mu(\pi)$ does not decay so steeply on $I_L$ as to overcome the $(1-\pi)^3$ advantage. In our numerical illustrations (with $\mu\equiv\mu_0>0$ constant), (C1) holds trivially.

\paragraph{Consequence of (C1).}
By Eq.~\eqref{eq:app_C1_limit}, there exists $s_0>0$ such that for all $\bar\sigma_\rho^2\in(0,s_0)$, $D(\bar\sigma_\rho^2)>\frac{C_{0}}{2}\bar\sigma_\rho^2>0$. In particular, picking any $s_a\in(0,s_0)$ gives $D(s_a)>0$.

\subsubsection{(C2) Existence of an off-policy regime where high-probability tokens dominate}
\label{app:C2_discussion}
Condition (C2) of Corollary~\ref{cor:gradient_dominance_reversal} asserts that there exists $s_1>0$ with $D(s_1)<0$. (C2) is not a consequence of the modeling choices alone --- one can construct degenerate $(\mu,p)$ pairs (e.g., $p$ supported entirely on $I_L$) for which $D(\bar\sigma_\rho^2)>0$ at every $\bar\sigma_\rho^2$. We discuss the mechanism that produces $D(s_1)<0$ for non-degenerate $(\mu,p)$, and how (C2) is verified from the closed form in our representative setup.

\paragraph{Mechanism.}
Eq.~\eqref{eq:app_closed_form} shows that the $\pi$-dependence of $\mathbb{E}[G\mid\pi]$ enters through $\tau^2(\pi)=\log(1+\sigma_\rho^2(\pi))$ and through the standardized window $[L'(\pi),U(\pi)]$. Recall from §\ref{app:derivation_proposition} that $H(\tau)$ peaks at an intermediate $\tau$ and decays for large $\tau$ (the trust-region indicator pushes mass into the tail of the tilted log-normal). Since $\sigma_\rho^2(\pi)$ is monotonically decreasing in $\pi$, $H(\tau(\pi))$ is maximized at small $\pi$ for small $\bar\sigma_\rho^2$ but for sufficiently large $\bar\sigma_\rho^2$ shifts past the peak on $I_L$ first, while $H(\tau(\pi))$ on $I_H$ stays closer to the small-$\tau$ regime $H(\tau)\approx\tau^2$. As $\bar\sigma_\rho^2$ grows, this mismatch reduces $\bar G_L$ relative to $\bar G_H$.

\paragraph{Numerical verification.}
Whether the mismatch is large enough to produce $D(s_1)<0$ depends on $(p,\mu,\epsilon_h,\epsilon_l)$. Given any concrete choice of these, $D$ is a one-dimensional function of $\bar\sigma_\rho^2$ (equivalently $\kappa$ via Eq.~\eqref{eq:app_sigmabar_kappa}) that can be evaluated by direct numerical integration of Eq.~\eqref{eq:app_closed_form} against $p(\pi)$. Using the representative setup of Appendix~\ref{app:numerical_setup} (i.e., $(\epsilon_h,\epsilon_l)=(0.2,0.2)$, $\mu(\pi)\equiv\mu_0$, and $p(\pi)$ from a Qwen2.5-7B rollout), we verify numerically that $D$ is positive for small $\bar\sigma_\rho^2$ and becomes negative at sufficiently large $\bar\sigma_\rho^2$, confirming (C2) for this representative case. The qualitative pattern is robust to mild changes in $p(\pi)$ and to slowly-varying $\mu(\pi)$.

\paragraph{Why the reversal is not exponential.}
Since $\tau^2(\pi)=\log\!\big(1+\sigma_\rho^2(\pi)\big)$ grows only logarithmically in $\sigma_\rho^2(\pi)$, the suppression of $H(\tau(\pi))$ on $I_L$ as $\bar\sigma_\rho^2\to\infty$ is polynomial in $\bar\sigma_\rho^2$, not exponential. Consequently, the reversal threshold $s^\star$ may sit at moderate values of $\bar\sigma_\rho^2$ rather than in an extreme tail; this is consistent with the empirical pattern in Figure~\ref{fig:empirical_validations_combined}(b), where dominance shifts occur over a moderate range of off-policy strengths.

\subsubsection{IVT crossing}
\label{app:IVT_crossing}
$D$ is continuous in $\bar\sigma_\rho^2\in(0,\infty)$: $\tau(\pi)$ is continuous in $\sigma_\rho^2(\pi)$ (and hence in $\bar\sigma_\rho^2$ via Eq.~\eqref{eq:app_sigmabar_kappa}) for each $\pi$, $H(\tau)$ is continuous in $\tau$, and the integrals defining $\bar G_L,\bar G_H$ are bounded (since $H$ is bounded on bounded $\tau$-ranges and $p(\pi),\mu(\pi),(1-\pi)$ are bounded on $[0,1]$), so dominated convergence ensures continuity.

By (C1), there exists $s_a\in(0,s_1)$ with $D(s_a)>0$.\footnote{Take any $s_a\in(0,\min(s_0,s_1))$ where $s_0$ is from §\ref{app:C1_proof}.} By (C2), $D(s_1)<0$. By the intermediate value theorem applied to $D$ on $[s_a,s_1]$, there exists at least one $s^\star\in(s_a,s_1)$ with
\begin{equation}
D(s^\star) \;=\; 0.
\end{equation}
We call any such $s^\star$ a \emph{gradient dominance reversal threshold}. When $D$ changes sign at $s^\star$ (the generic case), low-probability tokens dominate the conditional gradient signal in some left neighborhood of $s^\star$ and high-probability tokens dominate in some right neighborhood. Without further regularity, the IVT alone gives existence of a crossing but not uniqueness or monotonicity --- multiple crossings are not excluded by (C1)--(C2). We do not need uniqueness for the qualitative reversal claim made in the main text.

\subsubsection{Connection to the empirical observation}
The empirical pattern in Figure~\ref{fig:empirical_validations_combined}(b) is consistent with this model-level prediction. Two caveats: (i) the figure measures parameter-gradient norms aggregated over tokens, not the closed-form $D(\bar\sigma_\rho^2)$ directly; (ii) the model's $\bar\sigma_\rho^2$ summarizes off-policy strength, while the experimental degree of off-policy is controlled indirectly via the number of updates per rollout. The match should therefore be read as evidence that the qualitative dominance reversal predicted by the closed form is observable in practice, rather than as a quantitative validation of the closed form itself.

\section{Geometric Direction Analysis with Principal Angles}
\label{app:geom_details}

This appendix details the geometric tool used in Section~\ref{sec:framework} to quantify the \emph{directional} similarity between high-dimensional gradients. Our goal is to compare gradients in a way that is (i) robust to global rescaling, (ii) stable in very high dimensions, and (iii) compatible with the matrix-shaped structure of transformer layer gradients.

\subsection{Why subspace alignment instead of vector angles}
A naive approach is to flatten all parameters and compute a cosine similarity between two gradient vectors. However, for large models this is often dominated by noise and by the many low-energy directions. In addition, gradients in neural networks are typically \emph{low-rank structured}: most update energy concentrates in a relatively small set of directions (e.g., due to correlated features and shared activations). For these reasons, we compare the \emph{dominant gradient subspaces} rather than individual flattened vectors.

Concretely, given a layer parameter matrix (e.g., a linear projection weight) with gradient
\[
W \in \mathbb{R}^{m \times n},
\]
we treat $W$ as a linear operator and extract its top-$k$ right singular vectors as a compact representation of its dominant update directions.

\subsection{Subspace representation via SVD}
For each gradient matrix $W$, we compute the singular value decomposition (SVD):
\[
W = U \Sigma V^\top,
\]
where $U \in \mathbb{R}^{m \times m}$ and $V \in \mathbb{R}^{n \times n}$ are orthonormal matrices, and $\Sigma$ is diagonal with singular values in non-increasing order.

We define the rank-$k$ dominant \emph{right} singular subspace
\[
\mathcal{S}(W;k) \;=\; \mathrm{span}(V_k),
\]
where $V_k \in \mathbb{R}^{n \times k}$ contains the top-$k$ columns of $V$. Intuitively, $V_k$ captures the principal directions in the input-feature space along which the gradient operator varies most strongly.

\paragraph{Choice of right singular vectors.}
We use right singular vectors because for weight matrices mapping $\mathbb{R}^n \to \mathbb{R}^m$, $V_k$ corresponds to directions in the input dimension. Empirically we found $V_k$ yields stable comparisons across layers of similar shapes. (Using left singular vectors $U_k$ gives qualitatively similar conclusions for our use case.)

\paragraph{Choosing $k$.}
We set $k=128$ in all experiments unless otherwise noted. This value is large enough to capture most gradient energy for typical transformer projection matrices, while keeping computations tractable. In preliminary sweeps, we found the mean principal angle is qualitatively stable for $k$ in a broad range (e.g., 64--256), provided $k$ does not exceed the smaller matrix dimension.

\subsection{Principal angles between gradient subspaces}
Given two gradients $W^A$ and $W^B$ for the same layer (e.g., computed under two different update constructions), we form orthonormal bases for their dominant subspaces:
\[
Q^A \in \mathbb{R}^{n \times k}, \quad Q^B \in \mathbb{R}^{n \times k},
\]
where $Q^A = V_k(W^A)$ and $Q^B = V_k(W^B)$.

The $k$ principal angles $\{\theta_i\}_{i=1}^k$ between the two subspaces are defined by:
\[
\cos(\theta_i) \;=\; s_i,
\]
where $\{s_i\}_{i=1}^k$ are the singular values of the cross-Gram matrix
\[
M \;=\; (Q^A)^\top Q^B \in \mathbb{R}^{k \times k}.
\]
Equivalently, if $M = \tilde{U}\,\mathrm{diag}(s_1,\dots,s_k)\,\tilde{V}^\top$, then
\[
\theta_i = \arccos(s_i), \quad i=1,\dots,k.
\]

\paragraph{Interpretation.}
Each $\theta_i \in [0^\circ, 90^\circ]$ measures how well the $i$-th most-alignable direction in one subspace matches the other. In particular:
\begin{itemize}
    \item $\theta_i = 0^\circ$ indicates a perfectly shared direction.
    \item $\theta_i = 90^\circ$ indicates orthogonality for that principal direction.
\end{itemize}
Small angles across many $i$ imply the two gradients share a similar dominant update subspace; large angles indicate a substantial rotation of dominant directions.

\subsection{Summary statistics reported}
The set of principal angles yields a spectrum of directional similarity. For compact reporting, we aggregate angles into summary statistics.

\paragraph{Mean principal angle.}
We report the mean principal angle:
\[
\bar{\theta}(W^A,W^B) \;=\; \frac{1}{k}\sum_{i=1}^k \theta_i.
\]
This provides a single scalar notion of subspace alignment. In our plots/tables, smaller $\bar{\theta}$ indicates stronger alignment.

\paragraph{Alternative aggregations (not required).}
Other useful summaries include the median, the maximum angle, or a weighted mean using singular values of $W$ as weights. We found the unweighted mean sufficient and stable for the comparisons in this paper.

\subsection{Numerical stability}
In exact arithmetic, all singular values of $M=(Q^A)^\top Q^B$ lie in $[0,1]$. In finite precision, slight violations (e.g., $1+10^{-7}$) can occur, which would make $\arccos(\cdot)$ undefined. We therefore clamp:
\[
s_i \leftarrow \min(1,\max(0,s_i)),
\]
before applying $\arccos$.

\subsection{Layer-wise computation and aggregation across layers}
We compute principal angles \emph{per layer} and (when presenting model-level summaries) aggregate across layers.

\paragraph{Per-layer procedure.}
For each layer weight matrix:
\begin{enumerate}
    \item Compute gradients $W^A$ and $W^B$ under two update constructions.
    \item Compute truncated SVDs to obtain $Q^A$ and $Q^B$.
    \item Form $M=(Q^A)^\top Q^B$, compute its singular values $\{s_i\}$, then angles $\{\theta_i\}$.
    \item Summarize with $\bar{\theta}$ (and optionally the full spectrum).
\end{enumerate}

\paragraph{Across-layer aggregation.}
To summarize directional similarity for a whole model, we average $\bar{\theta}$ over a set of layers. In the main experiments we include the main projection matrices (e.g., attention and MLP projections) and exclude bias vectors and layernorm parameters, for which matrix-shaped SVD-based comparisons are not meaningful.

\subsection{Practical notes}
\paragraph{Truncated SVD implementation.}
We use a truncated SVD to compute the top-$k$ singular vectors efficiently. For large matrices, randomized SVD yields significant speedups while maintaining sufficient accuracy for principal-angle estimates.

\paragraph{Why compare subspaces, not singular vectors directly.}
Singular vectors are defined up to sign flips and may swap under small perturbations when singular values are close. Subspace-based measures are invariant to such instabilities, making principal angles more reliable for comparing gradients from stochastic minibatches.

\subsection{Limitations}
Principal angles measure similarity of dominant \emph{linear} subspaces per layer. This is not a full characterization of optimization trajectories, and it does not capture higher-order geometry or parameter coupling across layers. Nevertheless, it is a simple, robust diagnostic for whether two update constructions point in broadly similar directions in parameter space.

\subsection{Reproducibility checklist}
For reproducibility, we specify:
\begin{itemize}
    \item Subspace dimension: $k=128$.
    \item Angle computation: SVD of $(Q^A)^\top Q^B$ with clamping to $[0,1]$.
    \item Layer selection: matrix-shaped parameters (projection weights); exclude biases and layernorm scalars.
    \item Reported statistic: mean principal angle $\bar{\theta}$ (and spectra in some ablations).
\end{itemize}

\section{Reconciling Heuristics: Probability, Entropy, and Clipping Dynamics}
\label{app:prob_vs_entropy}
The empirical contradiction is resolved by examining the interplay between probability, entropy, and the clipping mechanism. While a token's probability $\pi$ and the vocabulary's entropy $H$ are correlated, their formal relationship is bounded:

$$
\underbrace{- \pi \log \pi - (1-\pi) \log(1-\pi)}_{H_{\text{min}}(\pi)} \le H \le \underbrace{- \pi \log \pi - (1-\pi) \log\left(\frac{1-\pi}{V-1}\right)}_{H_{\text{max}}(\pi)}
$$

\begin{figure}[t]
    \centering
    \includegraphics[width=0.35\textwidth]{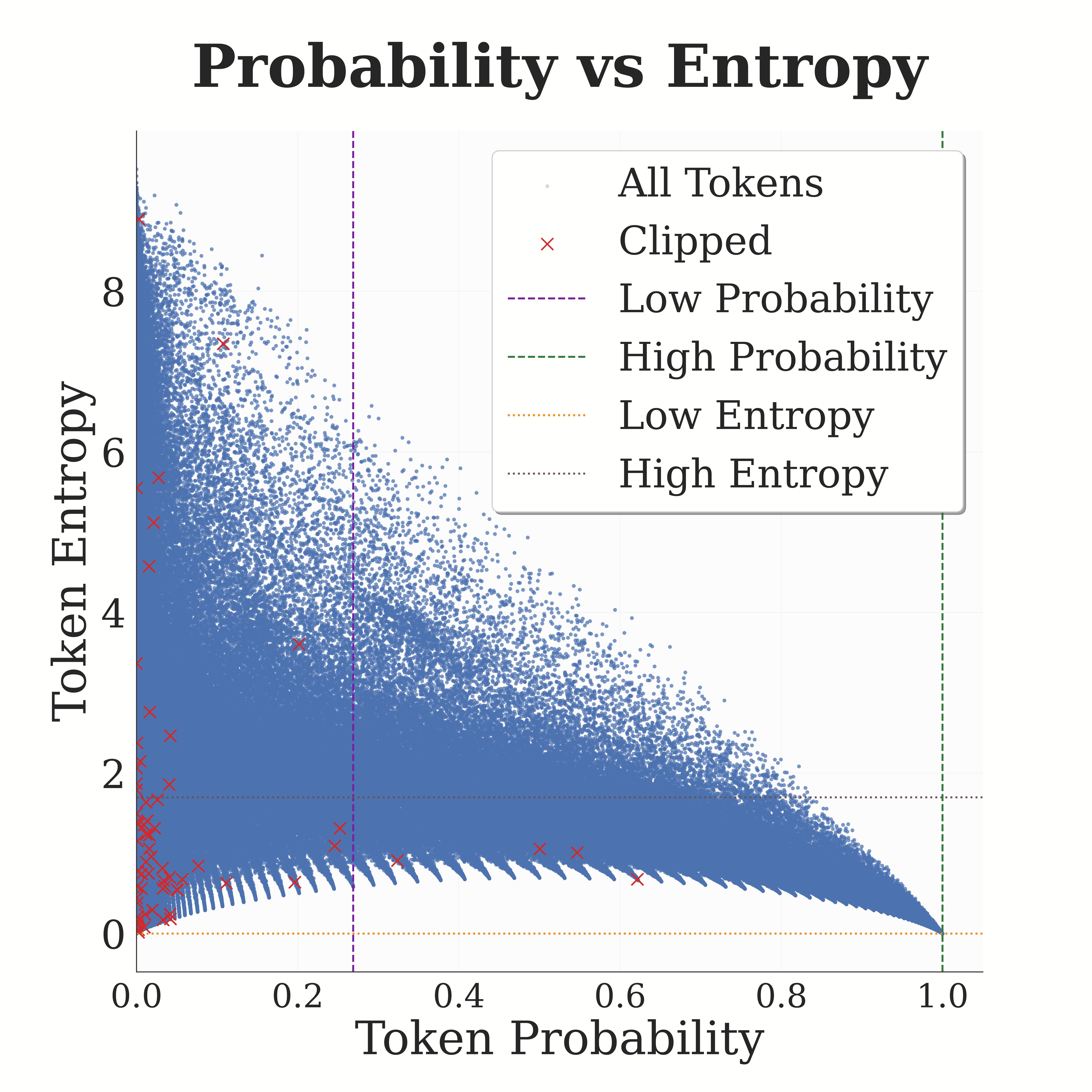}
    \caption{Correlation between token probability and entropy, with clipped tokens highlighted (red crosses).}
    \label{fig:prob_entropy_clipping_map}
\end{figure}

Figure~\ref{fig:prob_entropy_clipping_map} visualizes these bounds and provides the crucial insight: clipped tokens (red crosses) are not randomly distributed but are overwhelmingly concentrated in the low-probability, low-entropy region.

This single observation directly reconciles the conflicting findings:
\begin{itemize}
    \item Updating \textbf{low-probability tokens} \citep{yang2025not} is suboptimal because this population is dominated by dynamically unstable, low-entropy tokens. As established in Lemma~\ref{lemma:is_ratio_variance}, these tokens are highly susceptible to clipping, which attenuates their gradient contribution.
    \item Conversely, updating \textbf{high-entropy tokens} \citep{wang2025beyond} succeeds because it acts as an effective filter for stability. This heuristic implicitly avoids the heavily clipped region, selecting for tokens from more uniform distributions that are robust to policy updates.
\end{itemize}
Therefore, entropy is not merely a proxy for low probability but a more discerning indicator of a token's dynamic stability under the clipping pressures that govern off-policy optimization.

\section{Detailed Gradient Analysis}
\label{app:detailed_analysis}

\begin{figure}[htbp]
    \centering
    \begin{subfigure}{0.32\textwidth}
        \centering
        \includegraphics[width=\textwidth]{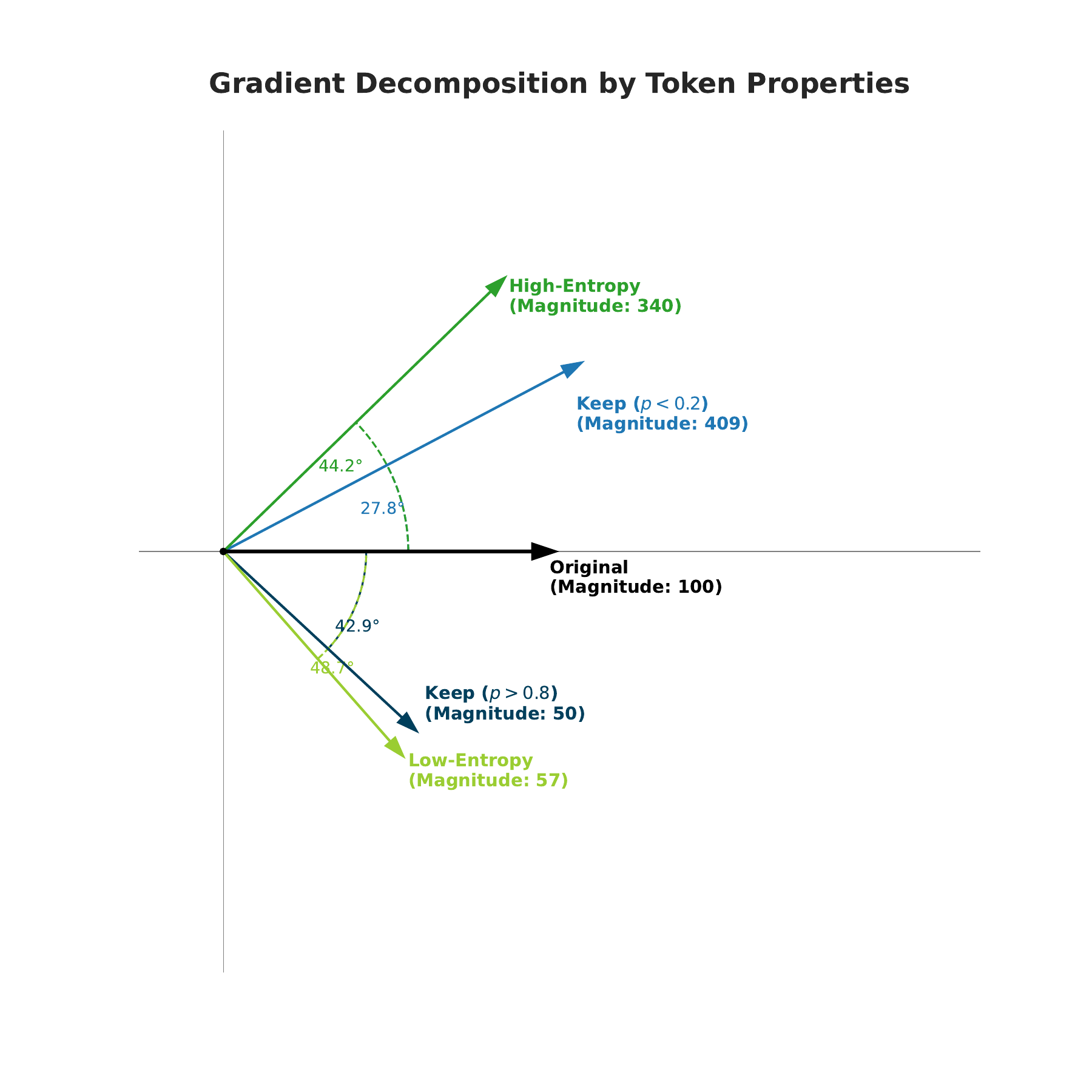}
        \label{fig:gradient_token}
    \end{subfigure}
    \hfill
    \begin{subfigure}{0.32\textwidth}
        \centering
        \includegraphics[width=\textwidth]{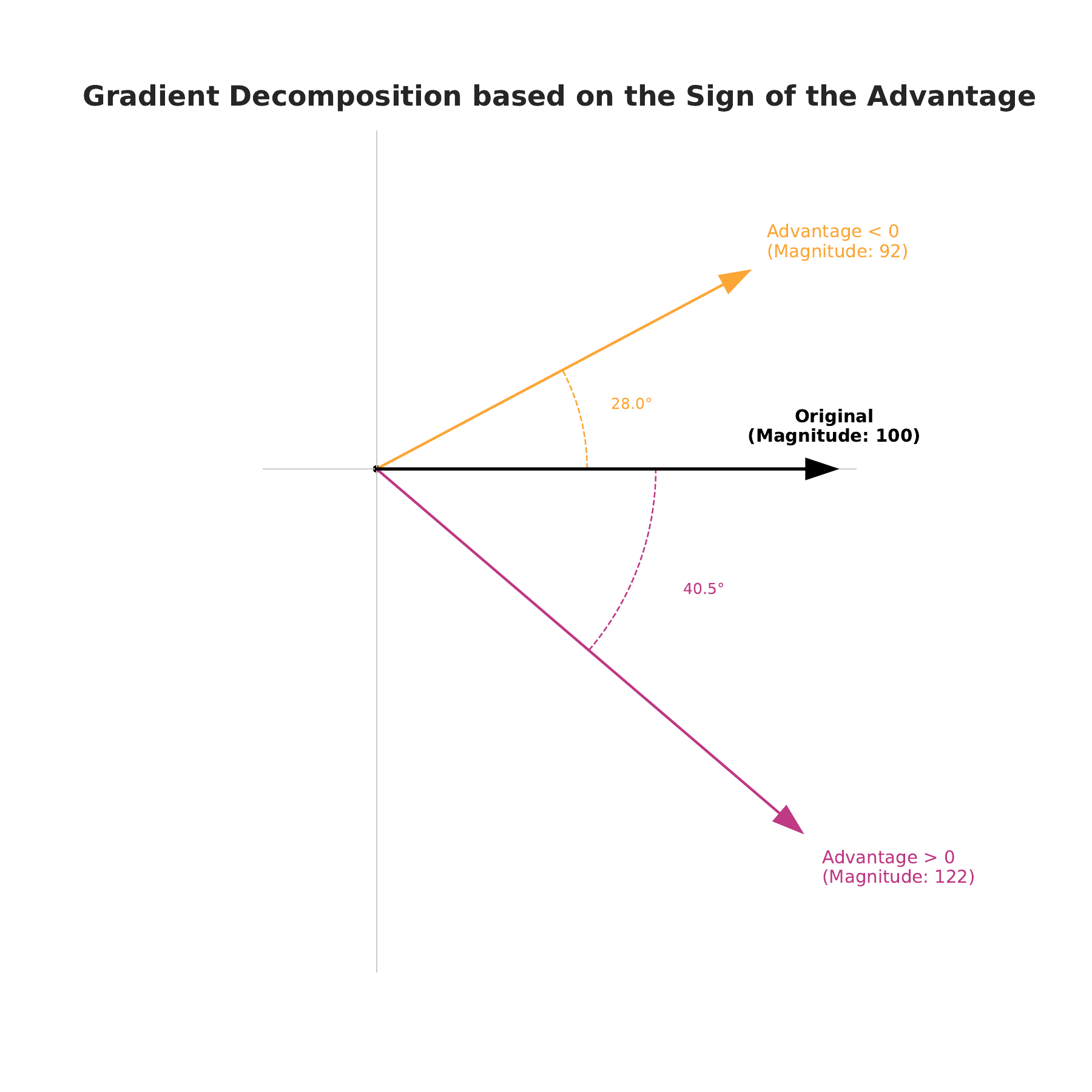}
        \label{fig:gradient_advantage}
    \end{subfigure}
    \hfill
    \begin{subfigure}{0.32\textwidth}
        \centering
        \includegraphics[width=\textwidth]{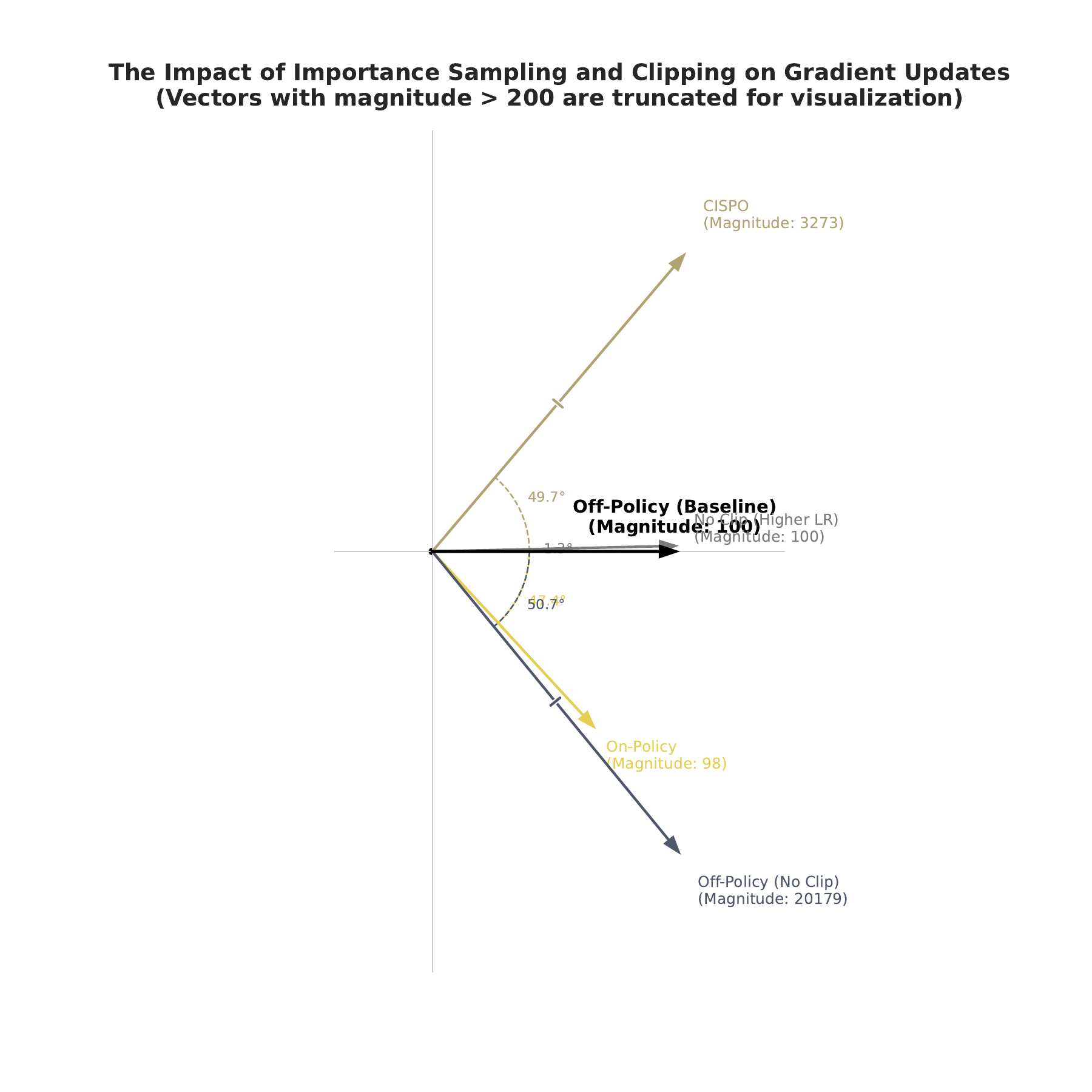}
        \label{fig:gradient_is}
    \end{subfigure}
    \caption{The impact of different key factors on RL updates.}
    \label{fig:key_factor}
\end{figure}

\subsection{Token Property Analysis}
\label{app:token_analysis}
To understand the fundamental composition of the policy gradient, we analyze the contributions of different token subsets based on their intrinsic properties. Our analysis, presented in Figure~\ref{fig:key_factor} left, isolates the gradient signals originating from distinct populations of tokens, revealing that the update is overwhelmingly dominated by a small, low-probability subset. Consistent with \citet{yang2025not}, tokens with $\pi_k < 0.2$ produce a gradient with a norm 409\% that of the full gradient and are exceptionally well-aligned with its direction (27.8$^\circ$ mean principal angle). In stark contrast, high-probability tokens ($\pi_k > 0.8$) contribute a much smaller, less aligned gradient (50\% norm, 42.9$^\circ$ angle). This confirms that the policy gradient is primarily driven by powerful but potentially unstable signals from the tail of the probability distribution.

The strategy of updating high-entropy tokens, proposed by \citet{wang2025beyond}, can be understood as an effective heuristic for managing this instability. The high-entropy subset produces a gradient with a substantial magnitude (340\% of original) but a weaker directional alignment (44.2$^\circ$) than the pure low-probability set. This suggests the high-entropy criterion functions as a filter: it selects for tokens that are informative (often having low to moderate probability) but originate from flatter, more uncertain distributions. As we will show, these flatter distributions are inherently more stable under policy updates. Therefore, the heuristic succeeds by implicitly balancing gradient magnitude against stability. This success underscores the need for a more principled mechanism that can dynamically manage this trade-off.

\subsection{Advantage Sign Analysis}
\label{app:advantage_analysis}
The advantage function, $A$, guides learning by signaling whether an action should be reinforced ($A>0$) or suppressed ($A<0$). To understand their distinct roles, we analyze the gradient contributions from these two subsets separately, as shown in Figure~\ref{fig:key_factor} middle. Our analysis reveals a critical asymmetry: while the gradient norms from positive (122\% of original) and negative (92\%) advantage samples are comparable, their directional guidance differs markedly. The gradient from negative-advantage samples is substantially better aligned with the total gradient direction (28.0$^\circ$ mean principal angle) than that from positive-advantage samples (40.5$^\circ$). This indicates that the overall update direction is predominantly dictated by corrective signals from suboptimal actions.

This directional dominance stems from the fundamentally different effects of positive and negative updates. An update with $A>0$ reinforces a single action, implicitly suppressing all alternatives and potentially narrowing the policy. Conversely, an update with $A<0$ penalizes a specific action, which effectively redistributes probability mass across the rest of the vocabulary. This latter mechanism, a form of error correction, promotes exploration and enhances policy diversity. The geometric dominance of negative-advantage gradients therefore suggests that learning in complex tasks is driven more by correcting errors than by reinforcing known correct pathways, a process crucial for discovering robust strategies. While \citet{zhu2025surprising} also noted the importance of negative samples, our work provides a novel perspective by demonstrating their dominant role in shaping the gradient's geometric direction.

\subsection{Importance Sampling Analysis}
\label{app:is_analysis}
To improve sample efficiency, off-policy reinforcement learning corrects for the policy distribution mismatch using Importance Sampling (IS). However, the high variance of the IS ratio necessitates a stabilization mechanism like clipping, which modulates the raw gradient into a stable update signal. To dissect this process, we compare our baseline off-policy GRPO update against several variants (Figure~\ref{fig:key_factor} right). The indispensability of clipping is starkly illustrated by its removal: the gradient norm explodes to \textbf{20,179\%} of the baseline, and its direction severely deviates (\textbf{50.7$^\circ$} mean principal angle). This confirms clipping is crucial not just for controlling magnitude but also for maintaining a stable optimization path.

Perhaps the most revealing finding is the significant directional divergence (\textbf{47.4$^\circ$}) between the on-policy and standard off-policy (GRPO) gradients, despite their nearly identical magnitudes (98\% vs. 100\%). This highlights a fundamental trade-off: to maintain magnitude stability, GRPO's fixed clipping mechanism systematically filters out certain updates (predominantly from high-variance, low-probability tokens), creating a gradient direction that is geometrically distinct from the on-policy ideal. This implies that while off-policy learning is sample-efficient, its optimization trajectory can substantially differ from its on-policy counterpart.

The specific design of the clipping strategy further modulates this trade-off. For instance, CISPO, which clamps the IS ratio instead of gating the entire gradient to zero, retains more signal from outlier tokens. This results in a substantially larger gradient norm (\textbf{3,273\%}) and a similarly large directional shift (\textbf{49.7$^\circ$}), reflecting a more aggressive update policy. In contrast, a simple asymmetric clipping scheme ($\epsilon_h > \epsilon_l$), inspired by DAPO, showed negligible impact in our setting (1.3$^\circ$ divergence). These comparisons reveal that the standard clipping in methods like PPO/GRPO is not a neutral stabilizer; it actively shapes the gradient by discarding certain information. This suggests an opportunity for adaptive mechanisms that can better preserve the on-policy direction while retaining off-policy efficiency.

\section{ACPO Formulation (Details)}
\label{app:acpo}

ACPO replaces the single global clipping range in Eq.~\ref{eq:GRPO} with a bin-specific range determined by the behavior-token probability.
Given an update mini-batch (pooling all tokens in the update batch), we assign each token \((i,t)\) to one of \(B\) equal-width bins on \([0,1]\) via
\begin{equation}
    b(i,t)=\min\{B,\lfloor B\cdot \pi_{\theta_{\mathrm{old}}}(o_{i,t}\mid q,o_{i,<t})\rfloor+1\}.
\end{equation}
For each bin \(b\), we compute the within-bin dispersion of IS ratios on this mini-batch and set the clipping range as
\begin{equation}
\begin{aligned}
    \sigma_b &= \mathrm{Std}\big(\{\rho_{i,t}: b(i,t)=b\}\big),\\
    \epsilon_b &= \epsilon_{\mathrm{base}}+\alpha\,\sigma_b.
\end{aligned}
\end{equation}
We then apply token-wise clipping by using \(\epsilon_{b(i,t)}\) in Eq.~\ref{eq:GRPO}. The resulting objective is
\begin{equation}
\small
\begin{aligned}
&J_{\text{ACPO}}(\theta) = \mathbb{E}_{q \sim P(Q),\, \{o_i\}_{i=1}^G \sim \pi_{\theta_{\text{old}}}(O|q)}
\frac{1}{G}\sum_{i=1}^{G}\frac{1}{|o_i|}\sum_{t=1}^{|o_i|} \\
&\min\Big[
\rho_{i,t}A_i,\;
\mathrm{clip}\big(\rho_{i,t}, 1-\epsilon_{b(i,t)}, 1+\epsilon_{b(i,t)}\big)A_i
\Big].
\end{aligned}
\label{eq:ACPO_appendix}
\end{equation}

\begin{algorithm}[t]
\caption{ACPO (bin-wise adaptive clipping)}
\label{app:acpo_alg}
\small
\begin{algorithmic}[1]
\STATE \textbf{Input:} behavior policy $\pi_{\theta_{\mathrm{old}}}$, current policy $\pi_\theta$, batch of rollouts, $B$, $\epsilon_{\mathrm{base}}$, $\alpha$.
\STATE Compute token-wise IS ratios $\rho_{i,t} = \frac{\pi_\theta(o_{i,t}\mid q,o_{i,<t})}{\pi_{\theta_{\mathrm{old}}}(o_{i,t}\mid q,o_{i,<t})}$.
\STATE Assign each token $(i,t)$ to a probability bin $b(i,t)=\min\{B,\lfloor B\cdot \pi_{\theta_{\mathrm{old}}}(o_{i,t}\mid q,o_{i,<t})\rfloor+1\}$.
\FOR{$b=1,\dots,B$}
\STATE $\sigma_b \leftarrow \mathrm{Std}(\{\rho_{i,t}: b(i,t)=b\})$.
\STATE $\epsilon_b \leftarrow \epsilon_{\mathrm{base}} + \alpha\,\sigma_b$.
\ENDFOR
\STATE Optimize Eq.~\ref{eq:ACPO_appendix} using token-wise clipping bounds $(1-\epsilon_{b(i,t)},\,1+\epsilon_{b(i,t)})$.
\end{algorithmic}
\end{algorithm}

\section{More Related Works}
\textbf{Reinforcement Learning for Large Language Models. }
Reinforcement learning has evolved from a tool for preference alignment to a key technique for enhancing reasoning capabilities in large language models. Initially pioneered through Reinforcement Learning from Human Feedback (RLHF), RL methods like PPO~\citep{schulman2017proximal} were used to align models with human preferences using human-annotated data~\citep{ouyang2022training}. The landscape shifted dramatically with the emergence of RL with verifiable rewards (RLVR), which leverages objective, automatically verifiable feedback signals instead of subjective human preferences~\citep{lambert2024tulu3}. OpenAI's o1 model~\citep{jaech2024openai} first showcased that RLVR can effectively incentivize reasoning at scale, particularly in tasks like mathematics and programming. Building on this foundation, researchers developed improved algorithmic methods such as GRPO~\citep{shao2024deepseekmath} and its variants (e.g., DAPO~\citep{yu2025dapo}). Subsequent models trained with these methods, including DeepSeek-R1~\citep{guo2025deepseek}, QwQ~\citep{qwen2025qwq32b}, and AceReason-Nemotron~\citep{chen2025acereason}, demonstrate that strong reasoning capabilities can emerge through outcome-based optimization with online RL algorithms, establishing RLVR as a promising paradigm for developing reasoning-capable LLMs.

\textbf{Token-Level Dynamics and Update Strategies.}
While RLVR has demonstrated strong potential, its effectiveness is often determined by how token-level updates are applied during training. Recent research has explored multiple related directions: entropy- or probability-based methods that incorporate or reduce uncertainty~\citep{wang2025beyond,yang2025not,cheng2025reasoning,cui2025entropy,gao2025one, chen2025seed}, 
modifications to importance-sampling and clipping that stabilize gradients~\citep{roux2025tapered,chen2025minimax,su2025klear,zheng2025group,wang2025stabilizing}, 
advantage design strategies ranging from negative reinforcement to minimalist or segment-level credit assignment~\citep{zhu2025surprising,xiong2025minimalist}, 
and off-policy approaches that reshape update distributions~\citep{yan2025learning,ma2025learning,fu2025srft,arnal2025asymmetric}. However, studies in this area sometimes yield conflicting conclusions: for example, \citet{wang2025beyond} emphasize that high-entropy minority tokens drive effective learning, whereas \citet{yang2025not} show that low-probability(usually high-entropy) tokens can over-dominate gradients and should be suppressed. Such inconsistencies call for a deeper, component-wise analysis of the RL update pipeline. While recent work has started to examine how individual training techniques affect RL dynamics~\citep{liu2025part}, we still lack a comprehensive framework that can systematically explain how these components work together to drive effective policy updates in LLM reasoning.

\section{Training Details}
\label{app:detail_train}

We conduct experiments using the Qwen2.5-7B model on the ORZ dataset. All methods are evaluated under two settings: near on-policy (N-OnP) with a train batch size of 128 and PPO mini batch size of 64, and off-policy (OffP) with a train batch size of 256 and PPO mini batch size of 16. The key difference between these settings lies in the staleness of the rollout data: in the N-OnP setting, the policy is updated more frequently relative to the rollout data, while in the OffP setting, the policy undergoes more gradient steps before refreshing the rollout buffer.

All methods share the following common hyperparameters: a learning rate of 1e-6, a maximum prompt length of 1024 tokens, a maximum response length of 3072 tokens, and a rollout size $N=10$. We train for 15 epochs without KL regularization (KL coefficient = 0) and without entropy bonus (entropy coefficient = 0). The loss aggregation strategy is seq-mean-token-mean, which first averages over tokens within each sequence and then averages across sequences.

\paragraph{DAPO.} The baseline method uses asymmetric clipping with $\epsilon_{\text{low}}=0.2$ and $\epsilon_{\text{high}}=0.3$, without any additional modifications.

\paragraph{AR-Lopti.} This method combines Advantage Reweighting (AR) and Low-Probability Token Isolation (Lopti). The AR component reweights advantages using $\alpha_{\text{AR}}=0.3$, $\tau_{\text{AR}}=0.7$, and negative advantage weight $w_{\text{neg}}=1.0$. The Lopti component isolates low-probability tokens using a threshold $p_{\text{lopti}}=0.5$.

\paragraph{High-Entropy.} This method selectively updates only the tokens with the highest entropy. We set $k_{\text{ratio}}=0.2$, meaning only the top 20\% highest-entropy tokens are updated during training.

\paragraph{Low-Entropy.} Conversely, this method updates only the tokens with the lowest entropy. We set $k_{\text{ratio}}=0.8$, meaning only the bottom 80\% (i.e., lowest-entropy) tokens are updated.

\paragraph{ACPO.} Adaptive Clipping Policy Optimization dynamically adjusts the clipping range based on the within-bin importance-sampling ratio variance. We use $\alpha=3.0$ and $\epsilon_{\text{base}}=0.2$, with the adaptive clip range bounded between $\epsilon_{\text{min}}=0.0$ and $\epsilon_{\text{max}}=3.0$.

\paragraph{CISPO.} This method applies importance sampling-aware clipping with separate clip ratios for the importance sampling ratio: $\epsilon_{\text{is\_high}}=0.45$ and $\epsilon_{\text{is\_low}}=1.0$.

\begin{table}[t]
\centering
\caption{Common hyperparameters for Qwen2.5-7B experiments on MATH.}
\label{tab:common-hyperparameters}
\begin{tabular}{lcc}
\toprule
\textbf{Hyperparameter} & \textbf{N-OnP} & \textbf{OffP} \\
\midrule
Base Model & \multicolumn{2}{c}{Qwen2.5-7B} \\
Max Prompt Length & \multicolumn{2}{c}{1024} \\
Max Response Length & \multicolumn{2}{c}{3072} \\
Learning Rate & \multicolumn{2}{c}{1e-6} \\
Total Epochs & \multicolumn{2}{c}{15} \\
Rollout $N$ & \multicolumn{2}{c}{10} \\
Temperature & \multicolumn{2}{c}{1.0} \\
Validation $N$ & \multicolumn{2}{c}{8} \\
$\epsilon_{\text{low}}$ & \multicolumn{2}{c}{0.2} \\
$\epsilon_{\text{high}}$ & \multicolumn{2}{c}{0.3} \\
Entropy Coefficient & \multicolumn{2}{c}{0.0} \\
KL Coefficient & \multicolumn{2}{c}{0.0} \\
Loss Aggregation & \multicolumn{2}{c}{seq-mean-token-mean} \\
\midrule
Train Batch Size & 128 & 256 \\
PPO Mini Batch Size & 64 & 16 \\
\bottomrule
\end{tabular}
\end{table}

\begin{table}[t]
\centering
\caption{Method-specific hyperparameters.}
\label{tab:method-hyperparameters}
\begin{tabular}{ll}
\toprule
\textbf{Method} & \textbf{Specific Hyperparameters} \\
\midrule
DAPO & (Baseline, no additional parameters) \\
\midrule
AR-Lopti & $\alpha_{\text{AR}}$ = 0.3 \\
         & $\tau_{\text{AR}}$ = 0.7 \\
         & $w_{\text{neg}}$ = 1.0 \\
         & $p_{\text{lopti}}$ = 0.5 \\
\midrule
High-Entropy & entropy\_top\_k = True \\
             & $k_{\text{ratio}}$ = 0.2 \\
\midrule
Low-Entropy & entropy\_bottom\_k = True \\
            & $k_{\text{ratio}}$ = 0.8 \\
\midrule
ACPO & variance\_adaptive\_clip = True \\
     & $\alpha$ = 3.0 \\
     & $\epsilon_{\text{base}}$ = 0.2 \\
     & $\epsilon_{\text{min}}$ = 0.0 \\
     & $\epsilon_{\text{max}}$ = 3.0 \\
\midrule
CISPO & $\epsilon_{\text{is\_high}}$ = 0.45 \\
      & $\epsilon_{\text{is\_low}}$ = 1.0 \\
\bottomrule
\end{tabular}
\end{table}

\section{Ablation Study}
\subsection{Parameter Sensitivity}
\label{app:param_sensitivity}
We study the sensitivity of ACPO to its two main hyperparameters: the scaling factor $\alpha$ in the adaptive rule $\epsilon_b = \epsilon_{\mathrm{base}} + \alpha\,\sigma_b$, and the base clipping range $\epsilon_{\mathrm{base}}$.
Tables~\ref{tab:acpo_alpha_sensitivity} and~\ref{tab:acpo_epsilon_sensitivity} report results on ORZ-trained models, evaluated on in-domain and OOD math benchmarks, under both near on-policy (2 updates/rollout) and off-policy (16 updates/rollout) regimes.

Overall, ACPO is reasonably robust across a broad range of settings.
In particular, moderate values (e.g., $\alpha\in\{3,4\}$ with $\epsilon_{\mathrm{base}}\approx0.2$) achieve consistently strong average performance, while the per-benchmark optima can shift slightly between the on-policy and off-policy regimes, reflecting different clipping dynamics.
\begin{table}[t]
\centering
\caption{Sensitivity of ACPO to $\alpha$ with fixed $\epsilon_{\mathrm{base}}=0.2$ on ORZ-trained models, evaluated on multiple mathematical reasoning benchmarks, avg@8. Within each regime, the best score for each benchmark is \textbf{bolded}.}
\small
\setlength{\tabcolsep}{4pt}
\begin{tabular}{ccccccccc}
\toprule
$\alpha$ & $\epsilon_{\mathrm{base}}$ & Minerva & Math500 & AMC & AIME24 & AIME25 & Olympiad & AVG \\
\midrule
\multicolumn{9}{c}{\textbf{Near On-Policy (2 updates/rollout)}} \\
\midrule
2 & 0.2 & \textbf{31.23} & \textbf{79.85} & 53.92 & 20.00 & 12.50 & 42.66 & 40.03 \\
3 & 0.2 & 31.02 & 79.78 & \textbf{54.37} & 20.83 & \textbf{13.33} & 42.83 & \textbf{40.36} \\
4 & 0.2 & 30.64 & 79.63 & 53.31 & \textbf{21.25} & 12.92 & \textbf{43.17} & 40.15 \\
\midrule
\multicolumn{9}{c}{\textbf{Off-Policy (16 updates/rollout)}} \\
\midrule
2 & 0.2 & 31.46 & 80.35 & 50.90 & 15.42 & 10.00 & 41.67 & 38.30 \\
3 & 0.2 & \textbf{31.66} & \textbf{80.48} & \textbf{51.20} & 16.25 & 11.25 & 42.17 & 38.84 \\
4 & 0.2 & 31.51 & 80.43 & 50.66 & \textbf{16.67} & \textbf{12.08} & \textbf{42.41} & \textbf{38.96} \\
\bottomrule
\end{tabular}
\label{tab:acpo_alpha_sensitivity}
\end{table}
\begin{table}[t]
\centering
\caption{Sensitivity of ACPO to $\epsilon_{\mathrm{base}}$ with fixed $\alpha=3$ on ORZ-trained models, evaluated on multiple mathematical reasoning benchmarks, avg@8. Within each regime, the best score for each benchmark is \textbf{bolded}.}
\small
\setlength{\tabcolsep}{4pt}
\begin{tabular}{ccccccccc}
\toprule
$\alpha$ & $\epsilon_{\mathrm{base}}$ & Minerva & Math500 & AMC & AIME24 & AIME25 & Olympiad & AVG \\
\midrule
\multicolumn{9}{c}{\textbf{Near On-Policy (2 updates/rollout)}} \\
\midrule
3 & 0.0 & 29.87 & 77.95 & 48.04 & 19.58 & \textbf{13.75} & 41.20 & 38.40 \\
3 & 0.1 & 30.65 & 78.33 & 50.90 & 20.42 & 13.33 & 42.10 & 39.29 \\
3 & 0.2 & \textbf{31.02} & \textbf{79.78} & \textbf{54.37} & 20.83 & 13.33 & 42.83 & \textbf{40.36} \\
3 & 0.3 & 30.73 & 79.23 & 52.71 & \textbf{21.25} & 12.92 & \textbf{42.99} & 39.97 \\
\midrule
\multicolumn{9}{c}{\textbf{Off-Policy (16 updates/rollout)}} \\
\midrule
3 & 0.0 & 29.50 & 75.10 & 45.78 & 15.83 & 10.00 & 40.30 & 36.09 \\
3 & 0.1 & 30.71 & 77.73 & 48.49 & 16.25 & 10.83 & 41.33 & 37.56 \\
3 & 0.2 & \textbf{31.66} & \textbf{80.48} & \textbf{51.20} & 16.25 & 11.25 & 42.17 & \textbf{38.84} \\
3 & 0.3 & 31.12 & 78.88 & 50.15 & \textbf{16.67} & \textbf{11.67} & \textbf{42.36} & 38.48 \\
\bottomrule
\end{tabular}
\label{tab:acpo_epsilon_sensitivity}
\end{table}

\subsection{Effects of Clipping Thresholds}
\label{sec:ablation_clip_threshold}
\begin{table}[b]
\centering
\small
\caption{Ablation study on Countdown with Qwen2.5-3B-Instruct.}
\begin{tabular}{lcc}
\toprule
\textbf{Method / Setting} & \textbf{N-OnP.} & \textbf{OffP.} \\
\midrule
DAPO (Baseline)     & 73.27 & 74.38 \\
DAPO (Clip-High=Max)      & 74.65 & 75.86   \\
GRPO (Clip=Avg)      & 72.55    & 74.16 \\
\textbf{ACPO} & \textbf{75.74} & \textbf{76.27} \\
\bottomrule
\end{tabular}
\vspace{-1em}
\label{tab:countdown-ablation}
\end{table}

A natural question is whether ACPO's improvements stem from the adaptive mechanism itself, or simply from using different clipping threshold values. To disentangle these factors, we conduct ablation experiments on the Countdown task with Qwen2.5-3B-Instruct (Table~\ref{tab:countdown-ablation}), where we match fixed thresholds to ACPO's observed statistics:
(1) \textbf{GRPO (Clip=Avg)}: both upper and lower thresholds set to ACPO's weighted average clip range;
(2) \textbf{DAPO (Clip-High=Max)}: upper threshold set to ACPO's maximum observed clip range.

Under the \emph{avg} setting, the uniform threshold is dominated by high-probability tokens with low variance, making it more restrictive than standard GRPO. This overly tight bound suppresses valuable updates from low-probability tokens, reducing performance (72.55\% vs.\ 73.27\% baseline).
Under the \emph{max} setting, matching DAPO's upper bound to ACPO's maximum admits more signals and yields modest gains (74.65\% vs.\ 73.27\%). However, this uniform expansion ignores token-level variance: some tokens remain over-clipped while others become overly permissive, leaving performance below ACPO (75.74\%).

These results confirm that ACPO's benefit stems from adaptive calibration, rather than simply using larger or smaller threshold values.

\subsection{Sensitivity to Number of Bins}
\label{app:bin_sensitivity}
We study how ACPO depends on the number of probability bins $B$ used for adaptive thresholding. Table~\ref{tab:bin_sensitivity_math} reports results on ORZ-trained Qwen2.5-7B over six math benchmarks under both near on-policy and off-policy regimes, and Table~\ref{tab:bin_sensitivity_countdown} reports Countdown results on Qwen2.5-3B-Instruct.

\begin{table}[h]
\centering
\small
\setlength{\tabcolsep}{4pt}
\caption{Bin-number sensitivity of ACPO on ORZ-trained Qwen2.5-7B across math benchmarks.}
\label{tab:bin_sensitivity_math}
\begin{tabular}{llccccccc}
\toprule
Regime & Bins & Minerva & Math500 & AMC & AIME24 & AIME25 & Olympiad & Avg. \\
\midrule
\multirow{3}{*}{N-OnP.} & 4 & 30.93 & 78.60 & 52.56 & 19.58 & 12.08 & 41.86 & 39.27 \\
                       & 5 & 31.02 & 79.78 & 54.37 & 16.77 & 14.37 & 42.83 & 39.86 \\
                       & 6 & 30.58 & 78.88 & 53.92 & 17.92 & 14.17 & 41.33 & 39.47 \\
\midrule
\multirow{3}{*}{OffP.}  & 4 & 30.65 & 80.28 & 50.75 & 17.08 & 10.42 & 40.64 & 38.30 \\
                       & 5 & 31.66 & 80.48 & 51.20 & 16.35 & 10.10 & 42.17 & 38.66 \\
                       & 6 & 30.10 & 79.48 & 51.81 & 16.67 &  9.17 & 42.43 & 38.28 \\
\bottomrule
\end{tabular}
\end{table}

\begin{table}[h]
\centering
\small
\caption{Bin-number sensitivity on Countdown (Qwen2.5-3B-Instruct).}
\label{tab:bin_sensitivity_countdown}
\begin{tabular}{lcc}
\toprule
Bins & Countdown (N-OnP.) & Countdown (OffP.) \\
\midrule
4 & 75.20 & 75.90 \\
5 & 75.74 & 76.27 \\
6 & 75.43 & 75.65 \\
\bottomrule
\end{tabular}
\end{table}

The default $B=5$ achieves the best average performance, while $B=4$ and $B=6$ remain very close in all settings, indicating that ACPO is robust to moderate changes in bin count.

\section{Clipping and Gradient Dynamics of ACPO}
\label{app:clip_dynamics}

\begin{figure}[t]
\centering
\includegraphics[width=\columnwidth]{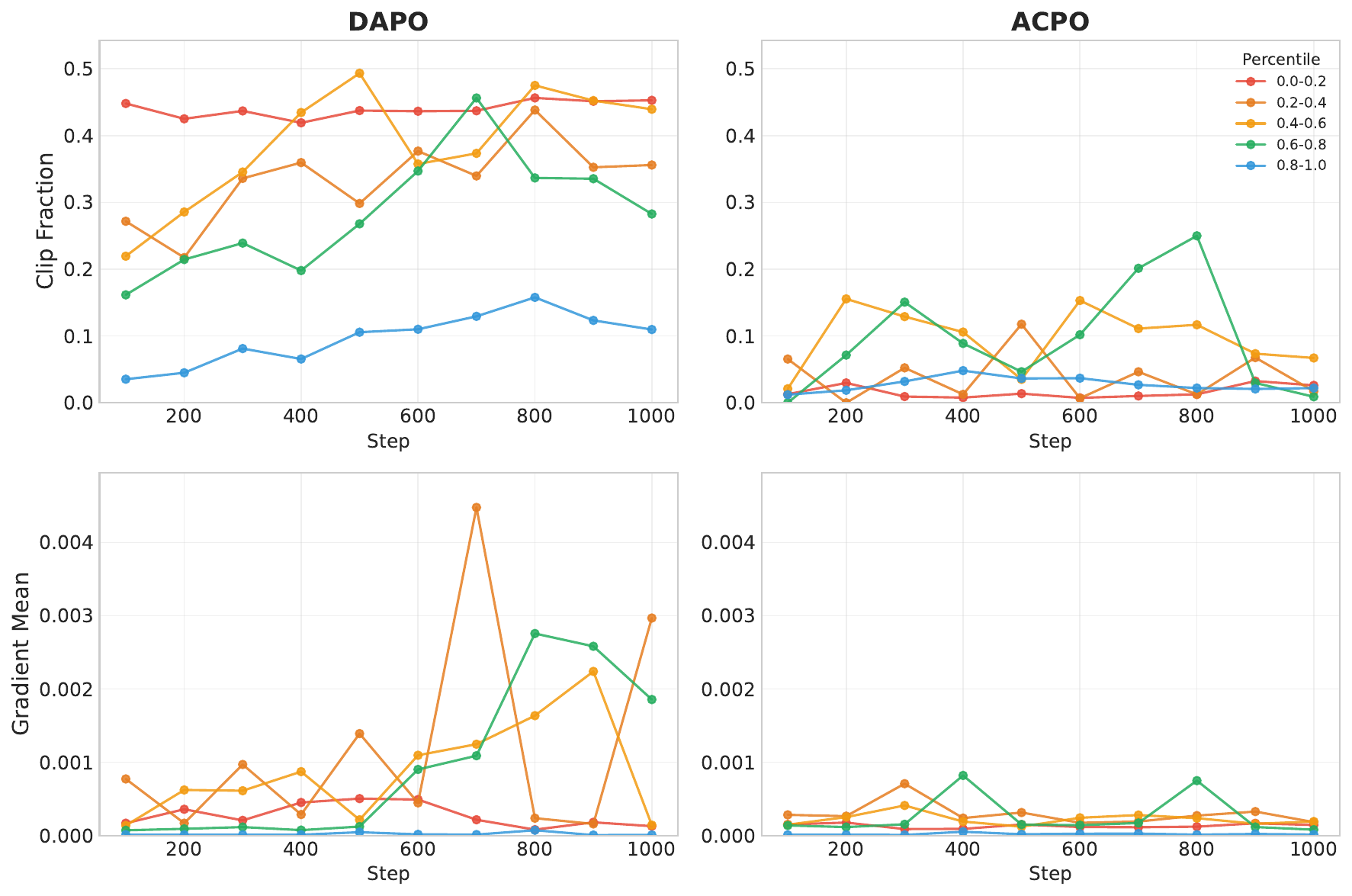}
\caption{Training dynamics under off-policy training across five probability bins. \textbf{Top:} Clip fraction over training steps. DAPO's fixed threshold yields heterogeneous rates; ACPO achieves uniform, low rates. \textbf{Bottom:} Gradient std over training steps. DAPO exhibits unstable spikes; ACPO remains stable throughout.}
\label{fig:training_dynamics}
\end{figure}

To understand the mechanisms underlying ACPO's stability, we analyze token-level clipping behavior and gradient dynamics during off-policy training on ORZ. Figure~\ref{fig:training_dynamics} compares DAPO and ACPO across five probability bins.

\textbf{Clip fraction.} DAPO exhibits heterogeneous clipping rates across bins: low-probability tokens (0.0--0.4) are clipped at 35--50\%, while high-probability tokens (0.8--1.0) at only 5--15\%. This disparity arises because fixed thresholds cannot accommodate the varying IS-ratio variance---tokens with inherently higher variance exceed the threshold more frequently, leading to over-clipping that discards potentially informative gradient signals. ACPO achieves more uniform clip rates across all bins ($<$20\%) by setting per-bin thresholds as $\epsilon_b = \epsilon_{\text{base}} + \alpha \cdot \sigma_b$, which naturally accommodates the heterogeneous variance structure.

\textbf{Gradient variance.} DAPO shows highly unstable gradient variance, with standard deviation spiking to over 0.02 for mid-probability bins (0.2--0.4) around step 600--800. ACPO maintains stable gradient variance below 0.005 throughout training for all bins, confirming that adaptive clipping provides consistent regularization across different token groups and training stages.

\section{Computational Overhead Analysis}
\label{sec:appendix-overhead}

\paragraph{Hardware and wall-clock cost.}
All training runs were carried out on a single node with 4$\times$NVIDIA A100 80GB GPUs. Under this configuration, a full RL training run takes approximately 14 days for the math reasoning task, 7 days for HiTab, and 2 days for Countdown. Reported numbers correspond to the experiments shown in the paper; preliminary and ablation runs consumed additional compute on the same hardware.

We analyze the computational overhead of different methods by measuring the per-token update time during training. Table~\ref{tab:update-time} summarizes the results across both Off-Policy and Near On-Policy settings.

\begin{table}[t]
\centering
\caption{Update time comparison. We report both per-step update time (seconds) and per-token update time (milliseconds). ACPO introduces negligible computational overhead compared to DAPO, while AR-Lopti incurs significant overhead due to additional token-level operations.}
\label{tab:update-time}
\resizebox{\textwidth}{!}{
\begin{tabular}{lcccccccc}
\toprule
\multirow{2}{*}{\textbf{Method}} & \multicolumn{4}{c}{\textbf{OffP}} & \multicolumn{4}{c}{\textbf{N-OnP}} \\
\cmidrule(lr){2-5} \cmidrule(lr){6-9}
 & Update (s) & Rel. & Per-token (ms) & Rel. & Update (s) & Rel. & Per-token (ms) & Rel. \\
\midrule
DAPO & 231.0 $\pm$ 31.7 & -- & 0.110 $\pm$ 0.005 & -- & 440.1 $\pm$ 48.9 & -- & 0.118 $\pm$ 0.003 & -- \\
CISPO & 239.6 $\pm$ 24.6 & +3.8\% & 0.128 $\pm$ 0.006 & +15.9\% & 485.1 $\pm$ 49.1 & +10.2\% & 0.136 $\pm$ 0.004 & +14.7\% \\
AR-Lopti & 359.4 $\pm$ 32.8 & +55.6\% & 0.234 $\pm$ 0.006 & +112.3\% & 659.9 $\pm$ 72.5 & +49.9\% & 0.258 $\pm$ 0.017 & +117.7\% \\
High-Entropy & 181.9 $\pm$ 22.9 & $-$21.2\% & 0.114 $\pm$ 0.005 & +3.3\% & 556.1 $\pm$ 107.6 & +26.4\% & 0.167 $\pm$ 0.028 & +40.6\% \\
\midrule
\textbf{ACPO (Ours)} & 219.1 $\pm$ 45.5 & $-$5.2\% & 0.110 $\pm$ 0.005 & +0.0\% & 458.4 $\pm$ 84.6 & +4.2\% & 0.109 $\pm$ 0.005 & $-$8.0\% \\
\bottomrule
\end{tabular}
}
\end{table}

\paragraph{Why ACPO has negligible overhead.}
ACPO (Adaptive Clipping Policy Optimization) adapts the clipping threshold based on the variance of importance-sampling ratios within each probability bin. The additional computation involves: (1) computing the within-bin standard deviation of IS ratios (a single reduction operation per bin), and (2) adjusting the clip ratio accordingly. Both operations are simple scalar computations performed once per batch, adding negligible overhead to the per-token update time.

\paragraph{Why AR-Lopti has significant overhead.}
AR-Lopti combines two techniques that both require additional per-token computations:
\begin{itemize}
    \item \textbf{Lopti (Low-Probability Token Isolation)}: Requires computing the probability of each token, comparing against a threshold ($p_{\text{lopti}}=0.5$), and creating a binary mask to isolate low-probability tokens. This involves additional tensor operations and conditional logic.
    \item \textbf{AR (Advantage Reweighting)}: Applies a non-linear reweighting function to advantages based on their signs and magnitudes, which requires additional computation beyond simple clipping.
\end{itemize}
Together, these operations result in over $2\times$ computational overhead compared to DAPO.

\paragraph{Why High-Entropy has moderate overhead.}
The High-Entropy method requires computing the entropy of each token's output distribution, sorting tokens by entropy, and selectively updating only the top-$k$ highest-entropy tokens. The entropy computation and sorting operations add approximately 40\% overhead.

In summary, ACPO achieves its performance improvements through computationally efficient per-bin adaptive clipping, making it practical for large-scale training without additional infrastructure requirements.

\section{Statistical Significance Tests}
\label{app:significance_tests}

We conduct statistical significance tests to compare ACPO against baseline methods on the OOD math benchmarks.
For each benchmark, we perform independent two-sample Welch's $t$-tests (unequal variances) using a one-tailed alternative hypothesis that ACPO outperforms the baseline: $H_1: \mu_{\text{ACPO}} > \mu_{\text{baseline}}$.

The test statistic is computed as:
\begin{equation}
t = \frac{\bar{X}_{\text{ACPO}} - \bar{X}_{\text{baseline}}}{\sqrt{\frac{s_{\text{ACPO}}^2}{n_{\text{ACPO}}} + \frac{s_{\text{baseline}}^2}{n_{\text{baseline}}}}},
\end{equation}
where $\bar{X}$ denotes the sample mean, $s$ the sample standard deviation, and $n$ the sample size.
The degrees of freedom are approximated using the Welch--Satterthwaite equation.
We use $n=8$ for Minerva, Math500, AMC, and Olympiad benchmarks, and $n=32$ for AIME24 and AIME25.
We consider results statistically significant when $p<0.05$.
\begin{table}[t]
\centering
\setlength{\tabcolsep}{5pt}
\caption{One-tailed Welch's $t$-test $p$-values comparing ACPO against each baseline on OOD math benchmarks, under near on-policy (2 updates/rollout) and off-policy (16 updates/rollout) settings. Bold indicates $p<0.05$.}
\label{tab:significance_tests}
\begin{tabular}{lcccccc}
\toprule
\multicolumn{7}{c}{\textbf{Near On-Policy (2 updates/rollout)}}\\
\midrule
ACPO vs. & Minerva & Math500 & AMC & AIME24 & AIME25 & Olympiad\\
\midrule
Base      & \textbf{$<0.001$} & \textbf{$<0.001$} & \textbf{$<0.001$} & \textbf{$<0.001$} & \textbf{$<0.001$} & \textbf{$<0.001$}\\
AR-Lopti  & \textbf{0.017}    & \textbf{$<0.001$} & \textbf{$<0.001$} & 0.052             & \textbf{$<0.001$} & \textbf{$<0.001$}\\
Top-20    & \textbf{0.018}    & \textbf{$<0.001$} & \textbf{0.029}    & \textbf{0.002}    & \textbf{$<0.001$} & \textbf{$<0.001$}\\
Bottom-80 & \textbf{0.007}    & \textbf{$<0.001$} & \textbf{$<0.001$} & 0.209             & \textbf{$<0.001$} & \textbf{$<0.001$}\\
DAPO      & 0.143             & \textbf{0.002}    & \textbf{0.007}    & --                & \textbf{0.005}    & \textbf{0.007}\\
CISPO     & 0.285             & \textbf{0.008}    & \textbf{$<0.001$} & --                & \textbf{$<0.001$} & \textbf{$<0.001$}\\
\midrule
\multicolumn{7}{c}{\textbf{Off-Policy (16 updates/rollout)}}\\
\midrule
ACPO vs. & Minerva & Math500 & AMC & AIME24 & AIME25 & Olympiad\\
\midrule
Base      & \textbf{$<0.001$} & \textbf{$<0.001$} & \textbf{$<0.001$} & \textbf{$<0.001$} & \textbf{$<0.001$} & \textbf{$<0.001$}\\
AR-Lopti  & \textbf{$<0.001$} & \textbf{$<0.001$} & \textbf{$<0.001$} & \textbf{$<0.001$} & \textbf{$<0.001$} & \textbf{$<0.001$}\\
Top-20    & \textbf{0.005}    & \textbf{$<0.001$} & \textbf{$<0.001$} & \textbf{0.013}    & --                & \textbf{$<0.001$}\\
Bottom-80 & \textbf{$<0.001$} & \textbf{$<0.001$} & \textbf{$<0.001$} & \textbf{$<0.001$} & \textbf{$<0.001$} & \textbf{$<0.001$}\\
DAPO      & \textbf{0.027}    & \textbf{$<0.001$} & \textbf{0.001}    & --                & \textbf{$<0.001$} & \textbf{$<0.001$}\\
CISPO     & \textbf{0.007}    & \textbf{$<0.001$} & \textbf{0.003}    & \textbf{0.002}    & \textbf{0.02}     & \textbf{$<0.001$}\\
\bottomrule
\end{tabular}
\end{table}

\section{Limitations}
\label{app:limitations}

\paragraph{Magnitude vs.\ direction analyses are not on equal theoretical footing.}
The magnitude side of the analysis in Section~\ref{sec:factors} is grounded in the closed form for $\mathbb{E}[G\mid\pi]$ derived in Section~\ref{sec:explain_the_difference}, so its conclusions about which token populations dominate the effective update are model-level statements with explicit assumptions. The directional side, by contrast, is empirical: principal angles between gradient subspaces are measured directly on training runs, without an accompanying closed-form characterization of high-dimensional parameter-gradient direction. We treat the directional findings as diagnostic evidence consistent with the magnitude-side mechanism, rather than as theoretical claims; closed-form modeling of update direction is beyond the scope of this paper.

\paragraph{Theoretical results rest on idealizations.}
The closed forms in Propositions~\ref{prop:gradient_expectation} and Corollary~\ref{cor:gradient_dominance_reversal} are derived under modeling choices stated in Section~\ref{sec:explain_the_difference} (log-normal IS ratio, linear advantage--log-ratio coupling, and the noiseless limit $\eta\equiv 0$). Robustness of the qualitative conclusions to the IS-ratio distribution is addressed in Appendix~\ref{app:gaussian_robustness} via a Gaussian re-derivation, and the finite-$\eta$ situation is acknowledged in Appendix~\ref{app:finite_eta}; we do not develop a full finite-$\eta$ closed form. Likewise, condition (C2) in Corollary~\ref{cor:gradient_dominance_reversal} is verified for the representative $(p,\mu)$ in Appendix~\ref{app:derivation_corollary} rather than for arbitrary distributions.

\paragraph{Empirical scope.}
Our experiments use the Qwen2.5 family (3B/3B-Instruct/7B) on math-reasoning and a small set of related tasks (ORZ, HiTab, Countdown). Cross-family generalization (e.g., to Llama-class or Mistral-class backbones) is not evaluated here, and we do not claim that ACPO's quantitative gains transfer unchanged across model families or task domains.

\paragraph{Scope: clipped, token-level IS.}
Our analysis and ACPO target the standard clipped, token-level IS setting (PPO/GRPO-style). The Section~\ref{sec:factors} results on token properties and advantage sign are policy-gradient statements that do not require clipping and would still apply to unclipped variants such as GPG~\citep{chu2025gpg}; in contrast, the IS-ratio variance, dominance reversal, and ACPO calibration in Sections~\ref{sec:analysis_contradictions} and \ref{sec:acpo} are specific to clipped objectives and do not directly carry over to unclipped methods. Likewise, our theory is for token-level IS; under sequence-level IS (e.g., GSPO~\citep{zheng2025group}, TIC-GRPO), the same dominance phenomenon would take a different form (which \emph{sequences}, rather than which tokens, survive clipping), and a separate analysis would be needed. Sequence-level IS is also better suited to settings such as MoE training and rollout/training mismatch, which we do not address here.



\end{document}